\pdfoutput=1
\documentclass[runningheads]{ks_press}
\usepackage[T1]{fontenc}
\usepackage{amsmath}
\usepackage{amsfonts}
\usepackage{amssymb}
\usepackage{hyperref}
\usepackage{graphicx}
\usepackage{romannum}
\usepackage{setspace}
\usepackage{booktabs} 
\usepackage[all]{nowidow}
\usepackage[utf8]{inputenc}
\usepackage{gensymb}
\usepackage{float}
\newcommand{\squeezeup}{\vspace{-6.5mm}}

\usepackage[mathlines]{lineno}
\usepackage{etoolbox} 
\newcommand*\linenomathpatchAMS[1]{%
  \expandafter\pretocmd\csname #1\endcsname {\linenomathAMS}{}{}%
  \expandafter\pretocmd\csname #1*\endcsname{\linenomathAMS}{}{}%
  \expandafter\apptocmd\csname end#1\endcsname {\endlinenomath}{}{}%
  \expandafter\apptocmd\csname end#1*\endcsname{\endlinenomath}{}{}%
}

\expandafter\ifx\linenomath\linenomathWithnumbers
  \let\linenomathAMS\linenomathWithnumbers
  
  \patchcmd\linenomathAMS{\advance\postdisplaypenalty\linenopenalty}{}{}{}
\else
  \let\linenomathAMS\linenomathNonumbers
\fi

\linenomathpatchAMS{gather}
\linenomathpatchAMS{multline}
\linenomathpatchAMS{align}
\linenomathpatchAMS{alignat}
\linenomathpatchAMS{flalign}

\makeatletter
\renewcommand\section{\@startsection{section}{1}{\z@}%
                       {-8\p@ \@plus -4\p@ \@minus -4\p@}%
                       {6\p@ \@plus 4\p@ \@minus 4\p@}%
                       {\normalfont\large\bfseries\boldmath
                        \rightskip=\z@ \@plus 8em\pretolerance=10000 }}
\renewcommand\subsection{\@startsection{subsection}{2}{\z@}%
                       {-8\p@ \@plus -4\p@ \@minus -4\p@}%
                       {6\p@ \@plus 4\p@ \@minus 4\p@}%
                       {\normalfont\normalsize\bfseries\boldmath
                        \rightskip=\z@ \@plus 8em\pretolerance=10000 }}
\renewcommand\subsubsection{\@startsection{subsubsection}{3}{\z@}%
                       {-4\p@ \@plus -4\p@ \@minus -4\p@}%
                       {-1.5em \@plus -0.22em \@minus -0.1em}%
                       {\normalfont\normalsize\bfseries\boldmath}}
\makeatother
\usepackage{graphicx}
\usepackage{capt-of}
\usepackage{makecell}
\usepackage{gensymb}
\usepackage{amsmath}
\usepackage{multirow}
\usepackage{adjustbox}
\usepackage[ruled,vlined]{algorithm2e}
\usepackage{setspace}
\graphicspath{ {./images/} }
\begin{document}
\setpagewiselinenumbers



%
\journal{XXXXX}
\volume{XXXXX}
\publisher{XXXXX}
\myDOI{XXXXX}


\title{Knowledge Graph Driven Approach to Represent Video Streams for Spatiotemporal Event Pattern Matching in Complex Event Processing}
\titlerunning{Video Event Knowledge Graph}
\author{Piyush Yadav \inst{1} \and Dhaval Salwala \inst{2} \and  Edward Curry \inst{1}}
\authorrunning{P.Yadav et al.}
\institute{Lero-Irish Software Research Institute, National University of Ireland Galway, Galway, Ireland\\
\email{\{piyush.yadav,edward.curry\}@lero.ie}
\and Insight Centre for Data Analytics, National University of Ireland Galway, Galway, Ireland\\
\email{dhaval.salwala@insight-centre.org}
}
\maketitle
\begin{history}
\received{(xx/xx/xx)}
\revised{(xx/xx/xx)}
\accepted{(xx/xx/xx)}
\end{history}

\begin{abstract}
Complex Event Processing (CEP) is an event processing paradigm to perform real-time analytics over streaming data and match high-level event patterns. Presently, CEP is limited to process structured data stream. Video streams are complicated due to their unstructured data model and limit CEP systems to perform matching over them. This work introduces a graph-based structure for continuous evolving video streams, which enables the CEP system to query complex video event patterns. We propose the Video Event Knowledge Graph (VEKG), a graph driven representation of video data. VEKG models video objects as nodes and their relationship interaction as edges over time and space. It creates a semantic knowledge representation of video data derived from the detection of high-level semantic concepts from the video using an ensemble of deep learning models. A CEP-based state optimization - VEKG-Time Aggregated Graph (VEKG-TAG) is proposed over VEKG representation for faster event detection. VEKG-TAG is a spatiotemporal graph aggregation method that provides a summarized view of the VEKG graph over a given time length. We defined a set of nine event pattern rules for two domains (Activity Recognition and Traffic Management), which act as a query and applied over VEKG graphs to discover complex event patterns. To show the efficacy of our approach, we performed extensive experiments over 801 video clips across 10 datasets. The proposed VEKG approach was compared with other state-of-the-art methods and was able to detect complex event patterns over videos with F-Score ranging from 0.44 to 0.90. In the given experiments, the optimized VEKG-TAG was able to reduce 99\% and 93\% of VEKG nodes and edges, respectively, with \textit{5.19X} faster search time, achieving sub-second median latency of 4-20 milliseconds.
\end{abstract}


\section{Introduction}
\label{S:1}
With the evolution of the Internet of Things (IoT), there is an exponential rise in sensor devices that are deployed ubiquitously. Due to the extensive usage of IoT applications in smart cities, smart homes, self-driving cars, and social media, there is humongous growth in multimedia data streams like videos and images. We are now transitioning to an era of Internet of Multimedia Things (IoMT)~\cite{alvi2015internet}, where visual sensors are pervasive. There is a significant shift in the data landscape where unstructured data like videos are continuously streamed from different IoMT devices like CCTV cameras and smartphones. Thus, the world is becoming more visual, where billions of images and videos are being uploaded and streamed every day. A recent report from Cisco~\cite{Cisco} estimated a 34\% Compound Annual Growth Rate (2017-2022) of consumer internet traffic is in the form of internet video streams. 

The \textit{middleware} acts as a communication abstraction between data producers (sensors) and consumers(applications), which are deployed in a distributed setting. Middleware systems like event-based systems enable consistent and timely event detection and mine patterns by analyzing the streaming data~\cite{hasan2015tackling}. Event processing systems are characterized by the concept of \textit{timeliness} which is collectively expressed with different terms like\textit{ on-the-fly, low-latency, high-throughput} and \textit{real-time processing}~\cite{hasan2015tackling,li2015supporting}.  Complex Event Processing is an event processing paradigm where streams of events are analyzed to gain high-level insights. CEP systems detect complex event patterns over streams using \textit{event rules}~\cite{cugola2012processing}. Various applications of CEP can be found in areas like environmental monitoring~\cite{broda2009sage}, energy~\cite{hill2008event}, and stock market analysis~\cite{demers2006towards}.

The CEP system performs pattern matching with an assumption that the incoming stream has some structured data model such as key-value pairs or RDF~\cite{cugola2015complex}. Currently, CEP systems have inherent limitations to process unstructured data streams. Unstructured data like videos come up with their own challenges as they are highly expressive to humans but lack a specific data model. For example, an image can be interpreted in multiple ways depending on human perception, have a high degree of information (multiple objects and relationships), but organized sparsely and are represented as sophisticated low-level features to the machine (pixels and edges). There is a requirement to structure such complex data to enable the CEP engine to reason and detect complex pattern matches in near real-time. This work is an extension of~\cite{Yadav2019} and focuses on defining a formal model and representation for video streams into high-level semantic concepts by correlating its low-level features. The defined structured model for the video stream is further optimized as per the CEP requirement for near real-time pattern matching.

The rest of the paper is organized as follows: Section 2 explains the initial background. Motivation, challenges, and problem requirements is described in Section 3. Section 4 explains the related work and the gap in the literature in video event matching in different domains. Section 5 throws light on defining video events and patterns in the CEP scenario. Section 6 enlists the video event representation approach, while Section 7 describes CEP based optimization over video representation for fast event matching. The system architecture is explained in Section 8, while Section 9 and 10 describe methods to create spatiotemporal relations and video event pattern rules in different domains. The experimental evaluation is explained in Section 11, and the paper concludes in Section 12.

\section{Preliminaries }
\label{S:2}

\squeezeup
\begin{table}[H]
\caption{Comparison of structured and unstructured data.}
\begin{center}
\begin{adjustbox}{width=9.2cm, height=2.4cm,center}
\begin{tabular}{| l | r | r | r |}
    \hline
    \makecell{\textbf{Characteristics}} & \makecell{\textbf{Structured Data}}   &
    \makecell{\textbf{Unstructured Data}} \\ \hline \hline
    \makecell{\textbf{Data Model}}  & \makecell{Predefined data \\ model and schema} & \makecell{No} \\ \hline
    \makecell{\textbf{Searchability}} & \makecell{Easy} & \makecell{Difficult} \\ \hline
    \makecell{\textbf{Ease of Analysis}} & \makecell{Easy} & \makecell{Difficult} \\ \hline
    \makecell{\textbf{Storage size}} & \makecell{Less} & \makecell{More} \\ \hline
    \makecell{\textbf{Interpretation}} & \makecell{Precise, ideal for databases \\ and easy fit for \\ machine interpretation} & \makecell{Multiple, depends on\\ human perception} \\ \hline
    \makecell{\textbf{Expressivity}} & \makecell{Precise} &\makecell{ Variable} \\ \hline
    \makecell{\textbf{Degree of Information} \\  Organization} & \makecell{High} & \makecell{Sparse} \\ \hline
    \makecell{\textbf{Examples}} & \makecell{RDF, XML, Key-Value} & \makecell{Images, Text, Video} \\ \hline
\end{tabular}
\end{adjustbox}
\end{center}
\label{tab:1}
\end{table}
\squeezeup

\subsection{Structured vs Unstructured Data}

Structured data have a fixed template and schema, while the unstructured data do not have any predefined schema or data model. The significant chunk of unstructured data consists of images, audio, and videos. A recent report from Oracle~\cite{Oracle} states that there is a 42.5\% growth of unstructured data like images and videos every year. Now, the internet constitutes nearly 80\% unstructured data making the web more visual. Table \ref{tab:1} enlists some significant differences between structured and unstructured data. 

\subsection{Complex Event Processing Systems}

\begin{figure}[H]
\centering
\includegraphics[width=9.1cm,height=3.3cm]{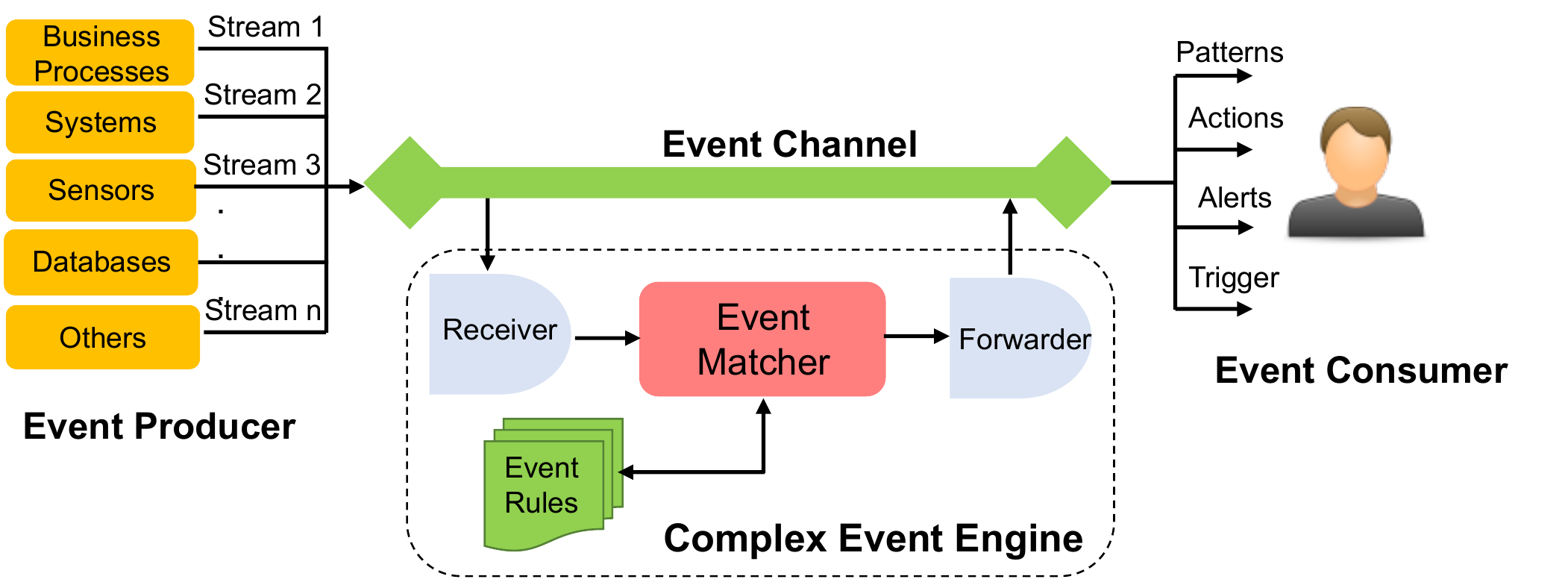}
\caption{Complex event processing paradigm}
\label{fig:1}
\end{figure}

Cugola and Margara~\cite{cugola2012processing} coined the term \textit{Information Flow Processing} (IFP) Systems, which originates from Active Databases~\cite{mccarthy1989architecture} and later followed in Data Stream Management~\cite{babcock2002models} and CEP~\cite{luckham2002power} systems. CEP system detects complex events by correlating simple events based on the registered event rule. The CEP matching model is continuous, where once the event rule is registered, the matching engine tries to mine patterns over incoming streams in an online setting. The system captures the recent \textit{state}~\cite{to2018survey} of the stream and applies a set of event rules and triggers notification as the pattern is detected. Figure \ref{fig:1} shows a high-level CEP architecture with different components.

\squeezeup
\squeezeup

\begin{itemize}
\item \textit{Event Producer}: The \textit{event producer} is the source which generates an event and sends it to the CEP system. There can be different types of producers like sensors, CCTV cameras, and databases that continuously collect data from the physical environment (figure \ref{fig:1}).
\item \textit{CEP Engine}: Figure \ref{fig:1} shows a high-level CEP engine component:
      \begin{itemize}
        \item{\textit{Receiver}: The \textit{receiver} receives the event from the \textit{event producer} through event channels and sends it to \textit{event matcher} for pattern matching.}
        \item{\textit{Event Matcher}: The \textit{event matcher} receives the input event and performs matching using \textit{event rules}. The matcher captures the recent \textit{state}~\cite{to2018survey} of the stream and applies a set of rules and triggers notification as the pattern is detected. Different matching models like automata-based, column-based~\cite{cugola2015complex}, and semantic matching~\cite{hasan2015tackling} have been proposed in the literature. The matched patterns are then sent to the \textit{forwarder}.}
        \item{\textit{Forwarder}: The forwarder receives the matched pattern and routes it to the \textit{event consumer} through event channel.}
      \end{itemize}
      
\item \textit{Event Consumer}: The event consumer can be a person, machine, or any other entity which receives the final matched pattern in the form of notifications, actions, and triggers.
\end{itemize}

\subsection{Knowledge Graphs}

\begin{figure}[H]
\centering\includegraphics[width=8.6cm,height=3cm]{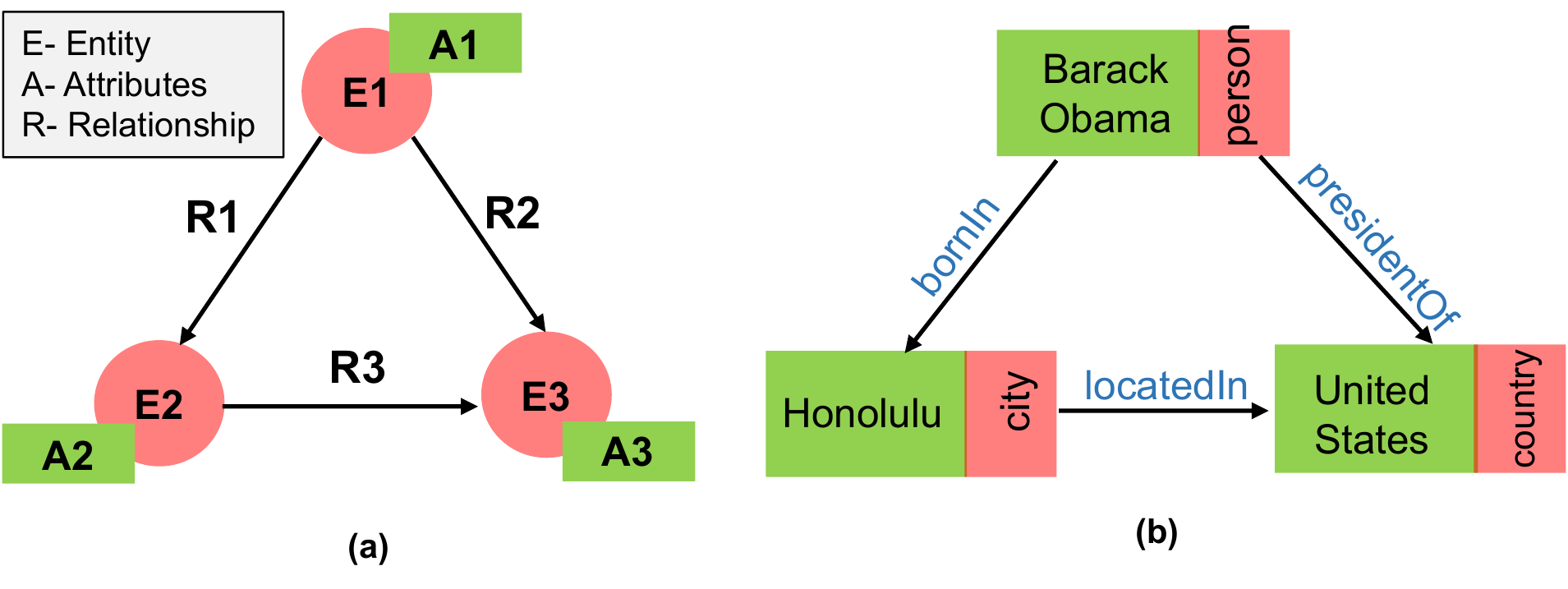}
\caption{(a) Knowledge Graph structure (b) Knowledge Graph example}
\label{fig:2}
\end{figure}

Knowledge Graph (KG) represents knowledge in graph form and captures entities, attributes, and their relation in nodes and edges respectively~\cite{van2008handbook}. Entities relate to things that exist in the real world and have an independent existence. Attributes are the characteristics and properties of an entity. A KG involves a two-step process- 1) construction and 2) query. The construction algorithm involves creating a graph from unstructured data by extracting entities, attributes, and relations using knowledge extraction and entity linking methods. The query algorithm then reasons over the constructed graph to identify patterns. Figure \ref{fig:2} (a) and (b) shows a KG structure with a simple example where person (E1) with name  (A1) Barack Obama was born in (R1) city (E2) Honulu (A2) and was the president of (R2) of country (E3) United States (A3). Here the edges (R1, R2, R3) are typed relationship with high-level semantic meaning like \textit{bornIn}, \textit{presidentOf}, and \textit{locatedIn}.  Some examples of famous KG’s are IBM Watson~\cite{IBM}, Google Knowledge Vault~\cite{dong2014knowledge}, and Facebook Graph API~\cite{Facebook}.

\squeezeup
\squeezeup

\subsection{ Image Processing and Understanding}

Video streams are continuous sequences of images. The image understanding domain focuses on reasoning over image content and describes the image using high-level semantic concepts. In computer vision, these high-level visual concepts are termed as \textit{objects}. There are various automated detection algorithms proposed in vision literature to detect objects from the images like SIFT~\cite{lowe1999object}. Recently, Deep Neural Networks (DNN)~\cite{lecun2015deep} has become a state-of-the-art method to detect objects with good accuracy and performance. A neural network is made of different connected layers (such as convolution, pooling, and ReLU), which learns high dimensional image features to perform the classification task. It is a supervised learning method where a model is trained using annotated training data to detect the presence or absence of an object in the given image. DNN based object detection models like YOLO~\cite{redmon2016you} and Faster R-CNN~\cite{ren2015faster} provide bounding boxes across the objects in the images which are highly accurate. DNN’s have opened a plethora of video analytics applications like monitoring garbage locations, crime prevention, and vehicle monitoring. Recent work like Scenegraphs~\cite{johnson2015image} tried to describe images by creating relationships between different objects to give more expressive representation to the images.

\section{Problem Formulation }
\label{S:3}

\subsection{Motivation}

\begin{figure}[H]
\centering\includegraphics[width=12cm,height=4.5cm]{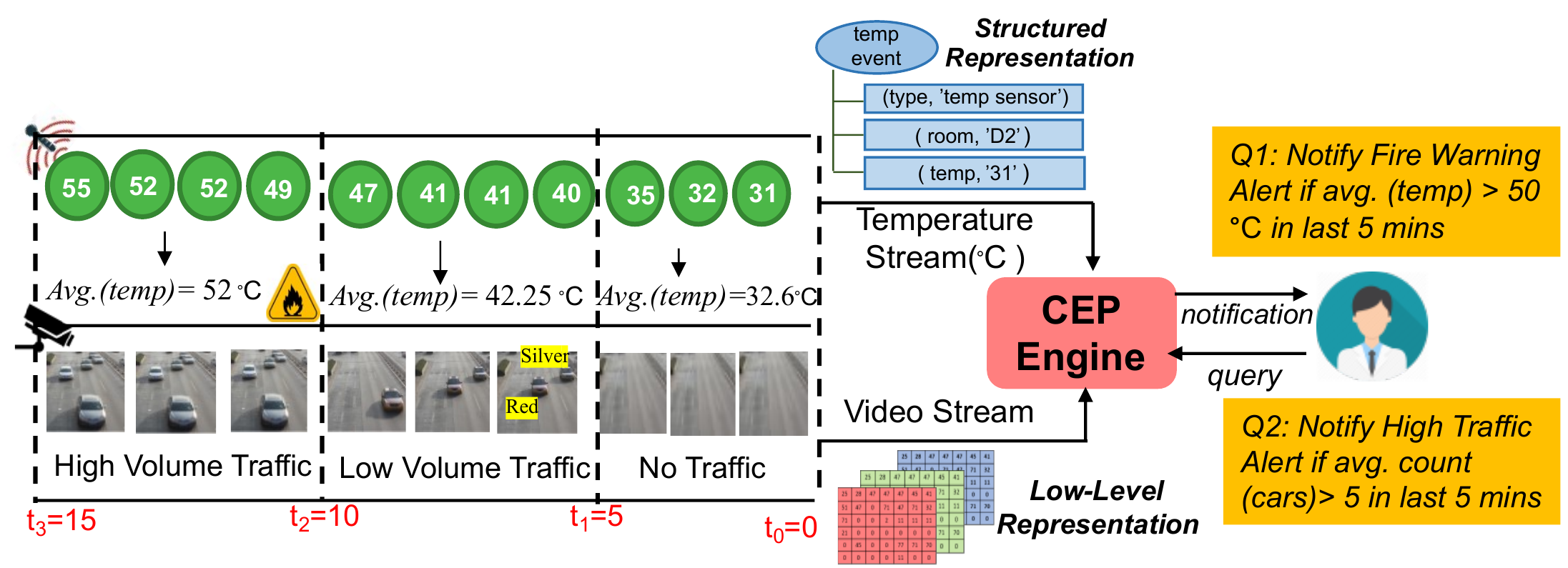}
\caption{Motivational Scenario}
\label{fig:3}
\end{figure}

Consider a smart city scenario where the city administrator has subscribed to the CEP system for a \textit{fire warning} and \textit{high volume traffic} alert. As shown in figure \ref{fig:3}, the CEP engine is receiving two data streams, one from a building temperature sensor and another from a CCTV camera. As per the query 1 rule (Q1), if the average temperature is greater than 50\degree C in the last five minutes, then the CEP system should notify a fire warning alert. The temperature sensor emits \textit{‘temperature event’} every second which is considered as a simple event. The complex event \textit{‘fire warning’} is derived by averaging \textit{‘temperature event’} for a given time. In figure \ref{fig:3}, a CEP system will raise a fire warning alert at time $t2-t3$ as the average temperature of incoming streams is higher than 50\degree C. Similarly, for query 2 (Q2), the CEP system should notify the traffic volume, but faces multiple challenges to process video streams. Most of the existing CEP and stream processing systems work with an assumption that the incoming stream has a structured format like key-value pairs (temperature = 50\degree C in figure \ref{fig:3}) and XML~\cite{cugola2015complex}. However, video data are complex and unstructured in terms of an event model. At the machine level, contents of the video data are represented as low-level features like color, pixels, shapes, and textures (figure \ref{fig:3}) while human interpret video content as a high-level semantic concept like car, chair, and person. While visualizing, human cognition can easily understand and differentiate events like \textit{no traffic} and \textit{high volume traffic}. It is difficult for CEP systems to reason over video data as 1) it has no structured representation and data model where semantic concepts boundaries are not known and organized, 2) the video event patterns spans over time and space.

\squeezeup
\squeezeup

\subsection{Video Event Detection in CEP: A Hybrid Approach }
\label{S:4}

Deploying different machine learning models for event detection in CEP scenario is challenging because of the following reasons:

\begin{itemize}
\item \textit{Huge Visual Concepts Space}: There are millions of visual concepts that are understood by humans. These concepts can be simple such as ‘car’ and complex, like ‘high volume traffic.’ With such large visual concept space, it is challenging to deploy machine learning models that can cover different visual concepts
\item \textit{Query Dynamicity}: Machine learning models like DNN are trained in supervised fashion to detect patterns. In CEP, there can be different continuous queries concerning users’ interest at a separate instance of time. It is not realistic to train every pattern where requirement changes due to subscriber’s query dynamicity.
\item \textit{Training Data Limitation}: The machine learning models’ performance is restricted with the amount of training data. The bigger the size of the training dataset, the better are the results. In the visual world, there are infinite objects and relationships, and it’s challenging to create a dataset for each pattern.
\end{itemize}

The training of the model for each pattern is costly in terms of resources and computation and infeasible in the CEP scenario. As discussed, the CEP system performs matching over pre-defined event rules. We propose a hybrid approach that includes inductive and deductive reasoning techniques. Inductive methods such as state-of-the-art DNN based models (like object detection and pose detection) can be used to detect initial simple events. Later, deductive reasoning methods (such as first-order logic)  can be applied to write event rules to create more complex event patterns using simple events. Following the hybrid approach, to process video stream queries (like Q2 in figure \ref{fig:3}) in a CEP engine leads to significant challenges:

\begin{itemize}
\item How to \textit{extract} and \textit{represent} low-level video content and video stream into a structured data model with high-level semantic concepts?
\item How to \textit{identify relationships} between semantic concepts of video content which occurs over time and space?
\item How to \textit{match} spatiotemporal CEP event rules over the represented data model \textit{efficiently} at run time?
\end{itemize}
To overcome the above challenges, basic requirements have been outlined which are required to model the video stream.

\subsection{Problem Requirements}
Videos are considered a continuous sequence of image frames, which consists of \textit{objects}. Humans perceive objects as a high-level semantic concept that occupy specific positions in an image. Technically, objects are a collection of low-level image features that have been given a high-level semantic label like \textit{car} and \textit{person}. Videos may have an evolving nature where different objects occur over time, generating varying kinds of complex events. Modeling complex events in unstructured data like videos require detecting objects and relationships between them. We enlist four essential requirements to represent video data suitable for complex event processing matching:
\begin{itemize}
    \item \textit{R1- Object Detection}: Objects are considered as fundamental building blocks of videos. There is a need to detect objects from low-level video content as they act as a backbone of the required data model. For example, a simple CEP query can be to notify if any \textit{car} object is present in the video feed.
\item \textit{R2- Attribute Detection}: An object can have specific characteristics that differentiate it from other objects. These can be termed as objects attributes. For example, in figure \ref{fig:3}, there are \textit{car} objects with color (red and silver) and type (sedan and van) attributes. 
\item \textit{R3- Spatiotemporal Relationship Identification}: Objects in videos are inextricably link with space and time as they change their spatial position (if moving) over time and generate different events.  These interactions create a spatiotemporal network giving rise to multiple complex events. For example, in figure \ref{fig:3}, a \textit{red car} is spatially located to the \textit{left} of a \textit{silver car} (frame 4). Here, \textit{left} is a spatial relationship between two \textit{car} objects with color attribute \textit{red} and \textit{silver}. Similarly, complex events such as \textit{high volume traffic} require the relationships across multiple objects.
\item \textit{R4: Event pattern matching}: The complex patterns can have both temporal and spatial events. These spatiotemporal factors need to be encoded in the representation by defining event rules. The event rules can be wrapped using a high-level semantic concept (such as high volume traffic), which can act as a template to match pattern over structured representation.  
\end{itemize}

\subsubsection{Contribution}

There is a clear need for a flexible video data model that can handle objects spatiotemporal dynamics. The main contribution of this work is as follows:
\begin{enumerate}

\item We present a flexible semantic event representation of video streams in the form of graphs using the Video Event Knowledge Graph (VEKG). The low-level video data is extracted and represented as a structured data model. VEKG model videos as streams of time-evolving graphs, where nodes and relationships change with time and space. VEKG acts as an intermediate bridge between unstructured video data and human level semantics.

\item A prototype architecture of the VEKG construction process. The architecture details all the components from video extraction to VEKG formation and captures explicit semantics of VEKG by modeling objects and their relationship interaction with each other.

\item A VEKG-Time Aggregated Graph (VEKG-TAG) which is an optimized VEKG representation for state-based CEP matching. VEKG-TAG is an \textit{aggregated representation} for VEKG over a given stream state. The VEKG-TAG is a storage efficient representation for modeling spatiotemporal networks like videos with 5.19X faster search with sub-second matching latency(4-20 milliseconds) and limited construction overhead.

\item We propose 9 event pattern rules using spatiotemporal calculus from two domains- 1) Activity Recognition, and 2) Traffic Management. The event rules are developed to show the efficacy of video pattern detection using VEKG in CEP environment.

\item An extensive experimental evaluation is performed over 801 videos across 10 datasets to show the overall performance of different CEP metrics and compared with the state-of-the-art methods. 

\end{enumerate}

\section{Related Work }
\label{S:5}

\subsection{Multimedia Event Representation}

Initially, Westermann et al.~\cite{westermann2007toward} proposed an ‘E’ event model for multimedia applications where they discussed various multimedia characteristics like spatial, temporal, informational, and structural, which need to be considered during event modeling. MSSN-Onto~\cite{angsuchotmetee2018mssn}  is an ontology framework which focuses on event schema for multimedia sensor network with visual descriptors, motion descriptor, spatial and temporal (camera duration) aspects instead of high-level semantic concepts and relationships in videos. IMGpedia~\cite{ferrada2017imgpedia} added low-level features of the image to create a linked dataset of images. The work is limited to static images instead of videos and captured no semantic relationship. In OVIS~\cite{tani2017ovis}, the authors have developed a video surveillance ontology using SWRL rules for large volumes of the video in databases. The OVIS focus was on indexing and retrieval of large size video in databases while VEKG representation can be deployed both, in the database and streaming scenario. Xu et al.~\cite{xu2015semantic} present Video Structural Description (VSD) technology for discovering semantic concepts in the video with no CEP focus. SPARQL-MM~\cite{kurz2015enabling} defines events in terms of spatial(point, line, shape) and temporal(instant, interval) thing for linked video. Works like Object Relation Network (ORN)~\cite{chen2012understanding} and CogVis~\cite{deshpande2017cogvis} propose a guide ontology to recognize the scene in an image while VEKG is used for modeling videos.
\subsection{Event Detection in Video Streams}

\subsubsection{Video Pattern Detection in Event Processing Systems} Gao et al.~\cite{gao2010spatio} focused on complex event detection in a multimedia communication system. They assumed the event as a high-level entity without any video content extraction being involved in the process. Taylor et al.~\cite{taylor2011ontology} proposed an event framework that uses ontology-driven CEP in sensor networks with structured digital messages and temporal correlations while VEKG deals with unstructured video data with both spatial and temporal relationships. Eventshop~\cite{singh2016situation} is a framework, which focuses on situation recognition in multimedia data where users model their situation of interest as per domain. The 'E-image' representation in Eventshop is limited and can not handle complexities like object's relative position, their presence and absence in a video with respect to time. VEKG representation is motivated from knowledge graphs, which can be enriched, queried, and can maintain the complex spatiotemporal relationship among different objects. Lee et al. proposed Region Adjacency Graphs (RAG)~\cite{lee2005strg} for videos where the same segmented regions within the image frames are connected using common boundaries. Instead of focusing on low-level features like in RAG, VEKG is built over high-level semantic labels (objects) extracted from DNN models capturing spatiotemporal relation among them. The work of Alam et al.~\cite{aslam2018towards} is limited to the detection of objects (like ‘Bus’) from images in event-based systems rather than focusing on complex event patterns. Yadav et al.~\cite{yadav2017event} focused on pattern detection like ‘wildfire’ from the unstructured event using crowd knowledge instead of automated detection of complex video patterns.

\subsubsection{Video Event Detection in Computer Vision}
Medioni et al.~\cite{medioni2001event} focused on detecting and tracking of moving objects and created a  frame-level representation using low-level image features. They do not deal with the formulation of a complex event pattern matching, query formulation over data streams, and stream representation. In ‘REMIND’~\cite{dubba2015learning}, Dubba et al. used Inductive Logic Programming to create relation event models for video. Our work overlaps with them regarding designing patterns but differs in the use of the CEP system, video event representation, and aggregation. Visual relation detection techniques like Scenegraph~\cite{johnson2015image} works on static data such as images where relationships are annotated among objects manually. VEKG, on the other hand, detects relationships among objects over space and time using event rules. Shang et al.~\cite{shang2017video} use neural networks and capture the relationship among objects in a video, where relationships were encoded in the training data manually and later trained to predict relation. In~\cite{herzig2018classifying}, Herzig et al. proposed a Spatio-Temporal Action Graph (STAG) to identify collision events using an end-to-end deep learning model. VEKG is a more generalized version of  STAG and can be used both in DNN and rule-based models and can handle multiple relationships within a single representation. Activity recognition~\cite{awad2016trecvid} is another domain that involves the detection of predefined human action like walk, jump, cook. Different models from First Order Logic rules such as Markov Logic Network~\cite{zhu2014reasoning} and Probabilistic Soft Logic~\cite{london2013collective} have been proposed to determine high-level activities.

\subsection{Graph Aggregation}
George et al. proposed TAG~\cite{george2008time}, an aggregated data model for spatiotemporal networks. VEKG-TAG is an addition to the above work, where we used it as an aggregation method over VEKG streams for a given time instance for detecting video event patterns. Kwon et al.~\cite{kwon2012unified} detect rare events in videos using a graph editing framework. They decompose video into a graph where a node represents a spatiotemporal event and have connected edges to its neighbors. In contrast, VEKG captures more detailed video information where each frame is initially a graph of objects with spatial information, which is then aggregated to VEKG-TAG over the temporal dimension for different queries. Adhikari et al. proposed NETCONDENSE~\cite{adhikari2017condensing}, which merges adjacent node-pair and time-pair for the time-varying graph. The time-pair merge loses initial edge information, which is preserved in the VEKG-TAG aggregation.

\section{Defining Video Events }
\label{S:6}

An event is an instantaneous occurrence of interest at a given point of time~\cite{luckham2002power} due to a change in the system environment. The video event is defined as a high-level semantic concept observed in the change of state in video content over time~\cite{francois2005verl}. The change in the video content can be due to multiple reasons such as object motion, new objects, the exit of old objects, and change in the appearance of objects over time. In CEP, the events can exist in entirety, i.e., simple events, or they can be derived from the collection of events, i.e., complex events. Following the  CEP terminology, two categories of video events are defined:

\begin{itemize}
\item \textit{Simple Video Event}: In CEP, a simple event is the instantaneous and atomic (i.e. either exists entirely or not at all) occurrence of interest at a specific time instance~\cite{wu2006high}. We have extended this notion of the simple event for videos. Objects are the primary visual concepts that a user can perceive from a video sequence.  Thus, a Simple Video Event can be defined as an occurrence of any object which a user can identify from the video. If a user queries about the presence or absence of objects (e.g. ‘car’, ‘person’) in a video, then we consider it as a Simple Video Event.

\item \textit{Complex Video Event}: In CEP, complex events are considered as \textit{composed} or \textit{derived} events that are constructed from simple events~\cite{demers2006towards}. The simple events are nested with different temporal and logical operators to form a complex event. Complex events can be either a collection of simple events or can be input to the formation of another complex event. The complexity of an event depends on the application logic and rules created for it. A Complex Video Event can be built using spatial, temporal, and logical operations using simple video events. For example, \textit{high volume traffic} in a video is a complex video event that is made from simple video events such as the presence of \textit{cars} and their count at a specific location for a given time.
\end{itemize}

\section{Video Event Representation }
\label{S:7}

\squeezeup
\begin{figure}[H]
\centering\includegraphics[width=6.2cm, height=4cm]{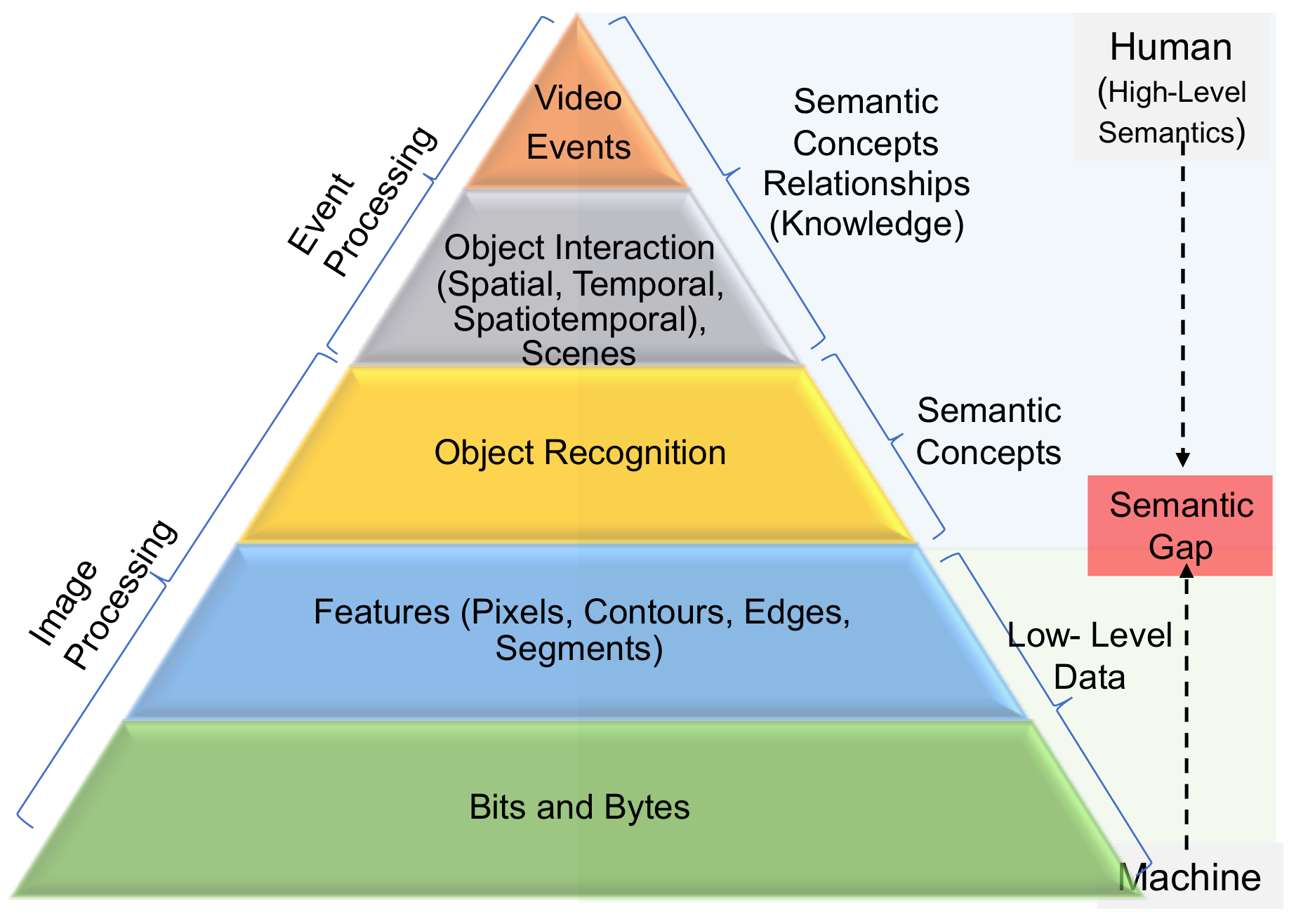}
\caption{Video events hierarchy and semantic gap between human and machine}
\label{fig:4}
\end{figure}
\squeezeup

In CEP, each incoming event is defined using a fixed data model. The stream is a continuous phenomenon, so each incoming data is timestamped. This helps in identifying their order of occurrences, which enables matcher for more complex reasoning during matching. Cugola et al.~\cite{cugola2015complex} represented an event in terms of \textit{payload} and \textit{time}. These payloads can have different structures and formats like key-value pairs structured XML and RDF triples.
Videos comprise a sequence of consecutive image frames and can be considered as a data stream, where each data item represents a single image frame. Representing semantic information from video streams is a challenging task. Figure \ref{fig:4} shows the complexity of a video event, where an image is represented as low-level features (pixel values) to the machine while human interprets them as high-level semantic concepts (‘car,’ ‘red car’) which creates a \textit{semantic gap} between the user and the data. In the real world, content extraction of video data leads to challenges like detecting object motions, relationships with other objects, and their attributes. Object detection techniques are not enough to define the complex relationships and interactions among objects and limit their semantic expressiveness. Thus, the video frames need to be converted into suitable representation to be processed by the CEP engine. We propose an object-centric representation using entity-centric Knowledge Graphs (KG). Graph-based representation for the video stream is suitable as it fits the following characteristics:

\begin{itemize}

\item \textit{Scalable}: Can capture multiple and diverse video objects and attributes information occurring at different time instances.
\item \textit{Complex Relationship}: Can capture interaction among video objects as spatiotemporal relationships which can later be inferred as a high-level event like high  volume traffic using event rules.
\item \textit{Maintains Hierarchy}: Can handle information at different hierarchies ranging from low-level image features to their semantic mappings like objects and scenes.
\item \textit{Semantically Queryable}: Can apply event rules and define pattern-matching operations over the data.

\end{itemize}

We have aligned the KG construction process with the video representation requirements (R1, R2, R3) listed in Section 3.3.  As shown in Figure \ref{fig:5}, the representation process is divided into two aspects- 1) Objects and Attribute Detection, and 2) Relationships among Objects.

\begin{itemize}

\item \textit{Objects and Attribute Detection (R1 \& R2)}: Following KG extraction, object and attribute detection is performed for video frames. The objects can have multiple characteristics and properties which are represented as their attributes (e.g., color, type). Figure \ref{fig:5} shows the extraction process for the image where car objects with different color attributes (red and black) are extracted.

\squeezeup
\begin{figure}[H]
\centering\includegraphics[width=9cm, height=3cm]{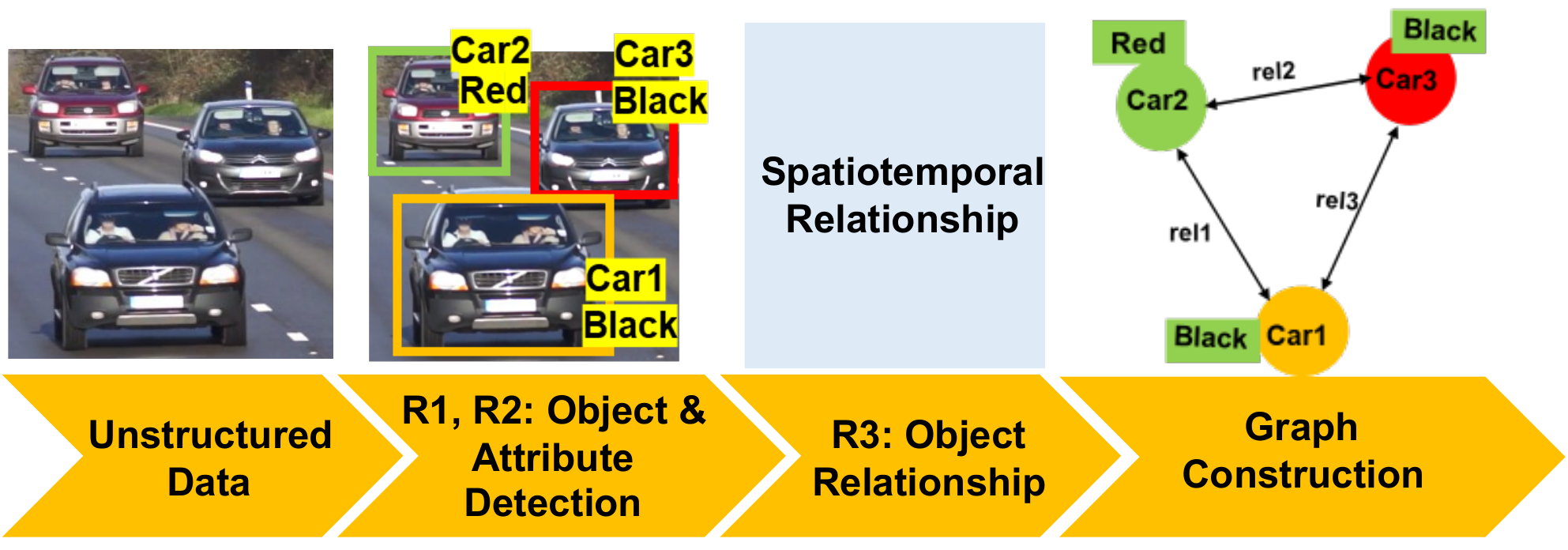}
\caption{VEKG extraction process~\cite{Yadav2019}}
\label{fig:5}
\end{figure}
\squeezeup
\item \textit{Relationships among Objects (R3)}: In a video, relationships among objects can exist across time and space. They can be classified as:

\subitem \textit{Relationship within a frame (Intraframe)}: Within an image frame, objects occupy specific positions. The objects may change their locations, creating multiple types of spatial interactions among different objects. Thus, a spatial relationship can be established among the objects within a frame. Figure \ref{fig:5} shows the spatial relation (rel1, rel2, rel3) among three car objects. The different spatial relationship will be discussed in Section 9.

\subitem \textit{Relationship across frames (Interframe)}: Across frames, objects interact with each other over time. The changeover object spatial interactions come with time. Thus, temporal relationships can be established among objects across frames. The temporal aspects among objects can be modeled as discrete or in the interval and will be discussed in Section 9.

\end{itemize}
Following this, a Video Event Knowledge Graph (VEKG) representation is proposed, where nodes correspond to objects and edges represent spatial and temporal relationships among objects (Figure \ref{fig:6}). A VEKG can be defined as:

\subsubsection{Definition1 (VEKG Graph):}For any image frame, the resulting Video Event Knowledge Graph is a labelled graph with six tuples represented as:
$VEKG=\{V,E, A_V, R_E,\lambda_V,\lambda_E\}$  where
\newline \mbox {V = set of object nodes } $O_i$
\newline \mbox {E = set of edges such } $E \subseteq V \times V$
\newline Av= set of properties mapped to each object nodes such that $O_i$= (id,attributes, 
\newline label, confidence,features)
\newline $R_E=$ set of spatiotemporal relations classes
\newline $\lambda_V,\lambda_E$ are class labeling functions- $\lambda_V \colon V \rightarrow O$ and $\lambda_E \colon E \rightarrow R_E.$

\subsubsection{Definition2 (VEKG Graph Stream):} A Video Event Knowledge Graph Stream is a sequence ordered representation of VEKG such that:
$VEKG(S)=\{(VEKG^1,t_1 ),(VEKG^2,t_2 ) \dots,(VEKG^n,t_n )\}$ where $t \in timestamp$   such that $t_i < t_{i+1}$. 

\squeezeup
\begin{figure}
\centering
\begin{minipage}{.5\linewidth}
  \centering
  \includegraphics[height=4cm, width=6.1cm]{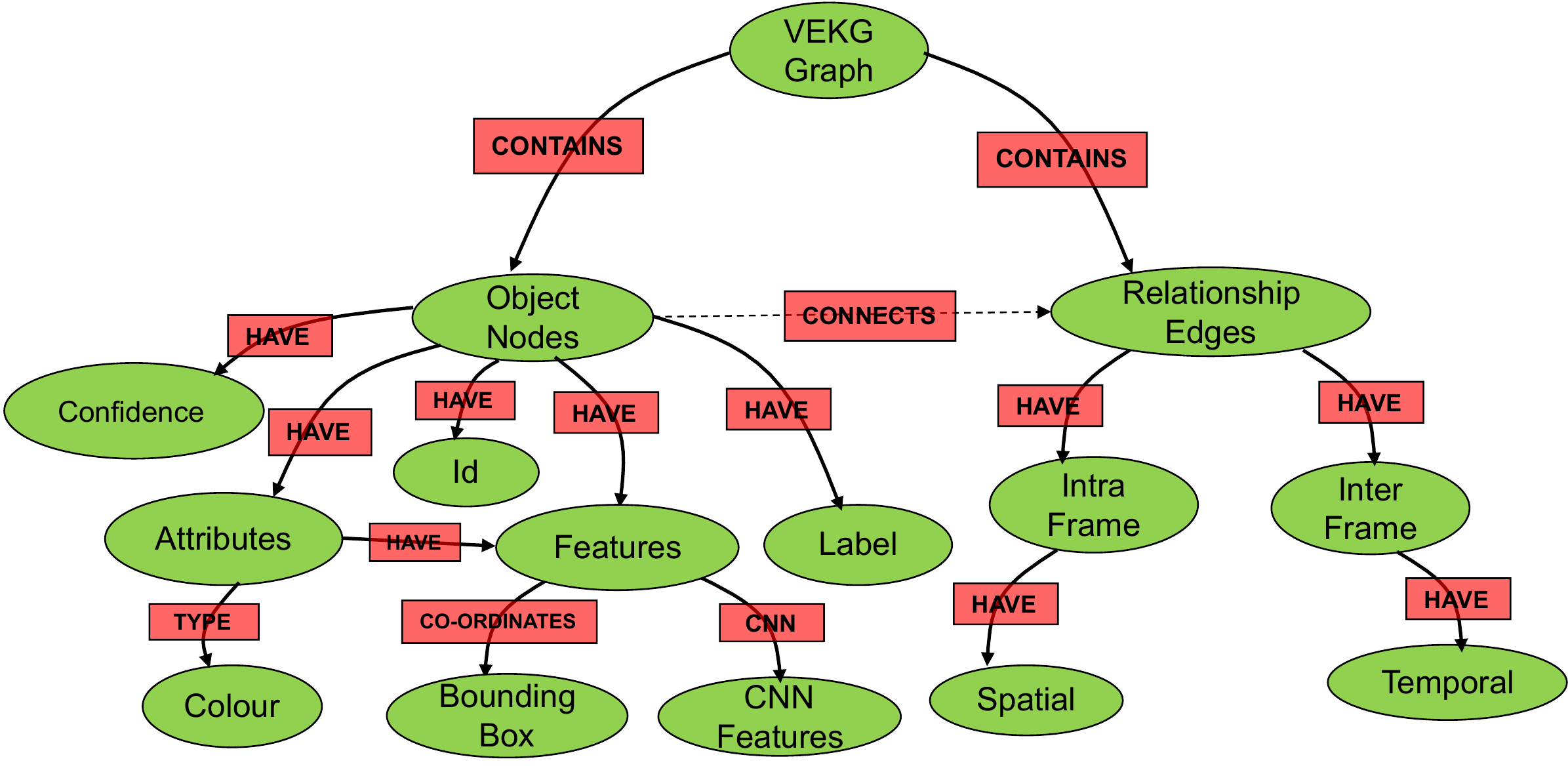}
  \captionof{figure}{Video Event Knowledge Graph (VEKG) schema~\cite{Yadav2019}}
  \label{fig:6}
\end{minipage}%
\begin{minipage}{.5\linewidth}
  \centering
  \includegraphics[height=4cm, width=6.1cm]{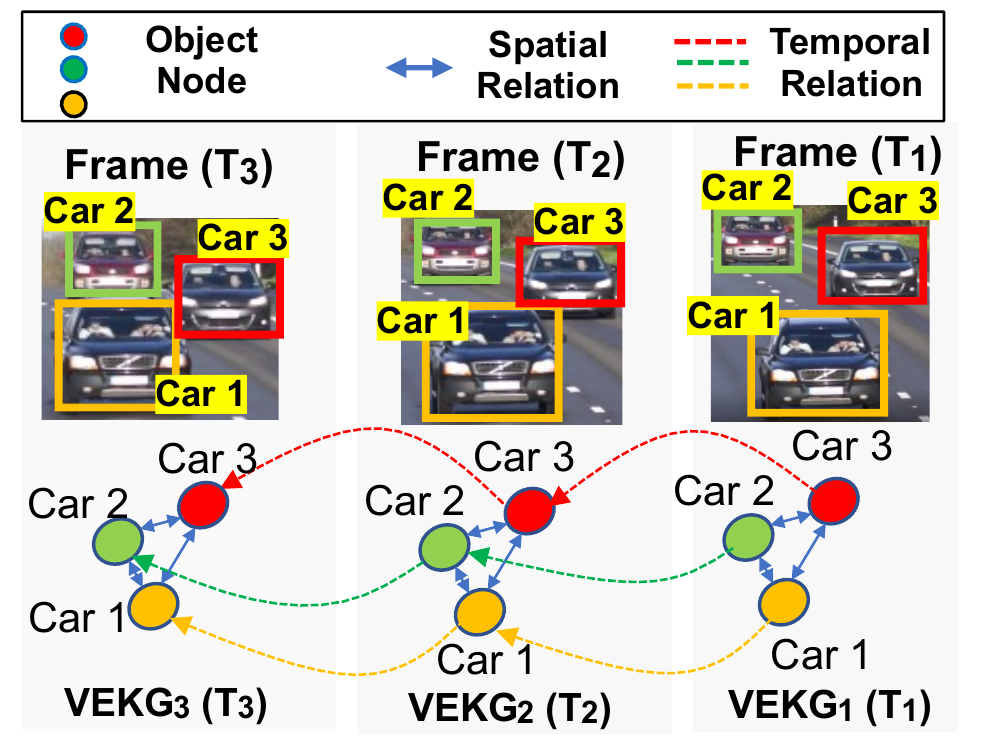}
  \captionof{figure}{VEKG graph stream~\cite{Yadav2019}}
  \label{fig:7}
\end{minipage}
\end{figure}
\squeezeup

Figure \ref{fig:7} shows the VEKG stream of video frames captured across three instances.  A \textit{directed graph} of spatial edges is created across object nodes to capture the spatial relationships among objects in the same frame. The object nodes (Car1, Car2, Car3) in VEKG graphs are connected using spatial edges. VEKG is a \textit{complete directed graph}, which means each object is spatially related to other objects which are present in the image frame. Thus each image frame consists of $n(n-1)$ edges where $n \in number \ of \ objects$. The edge weights between nodes are updated as per event rule and is discussed in Section 9. The temporal relationship edge between object nodes is created by identifying the same object nodes in different frames using object tracking.

\section{VEKG - Time Aggregated Graph: A state-based graph aggregation  }
\label{S:8}
Complex Event Processing(CEP) work on the concept of \textit{state} by discretizing continuous data streams in a fixed batch. The batches can be created using different metrics like \textit{time} and \textit{count}. Thus, the CEP matching for video streams can be divided into two types:
\begin{itemize}
\item \textit{Stateless Video Event Matching}: If the processing of an incoming event does not affect or depend on a subsequent incoming event stream, then it is considered as stateless video event processing. For example, detecting objects in every frame of video is considered as stateless matching.
\item     \textit{Stateful Video Event Processing}: The matching of patterns that requires a collection of events or where events get influenced by more than one input event is termed as stateful event matching. The complex patterns exist across time and space and require considerable event stream to detect patterns. For example, ‘high volume traffic’ patterns occur over space and time and requires multiple frames to analyze the patterns. In CEP, \textit{windows} capture these states and apply event rules over them to detect patterns. A window can be defined as:

\begin{equation}\scalebox{0.89}{$
TIMEWINDOW(VEKG(S),t) \colon \rightarrow S'
\label{eq:1}$}
\end{equation}
\end{itemize}

As per equation \ref{eq:1}, $TIMEWINDOW$ is applied over an incoming VEKG(S) stream and gives a  fixed subsequence $S^{'}$= $((VEKG^{1},t_{1} ) \dots, (VEKG^{n},t_{n}))$. In video, objects may exist for some time across multiple frames. As the objects are modeled as VEKG nodes, this leads to increase in the number of duplicate nodes which in turn increases the VEKG construction and search time. To reduce this overhead, we propose the Time Aggregated Graph (TAG)~\cite{george2008time} method over the VEKG stream. VEKG-TAG models time-series relationships across the edges of a single aggregated graph to accommodate the time-varying object interactions. VEKG-TAG gives an aggregated view of a video state for a given time interval that preserves all required relationships. VEKG-TAG can be defined as:

\squeezeup
\begin{figure}[H]
\centering\includegraphics[width=10cm, height=4.1cm]{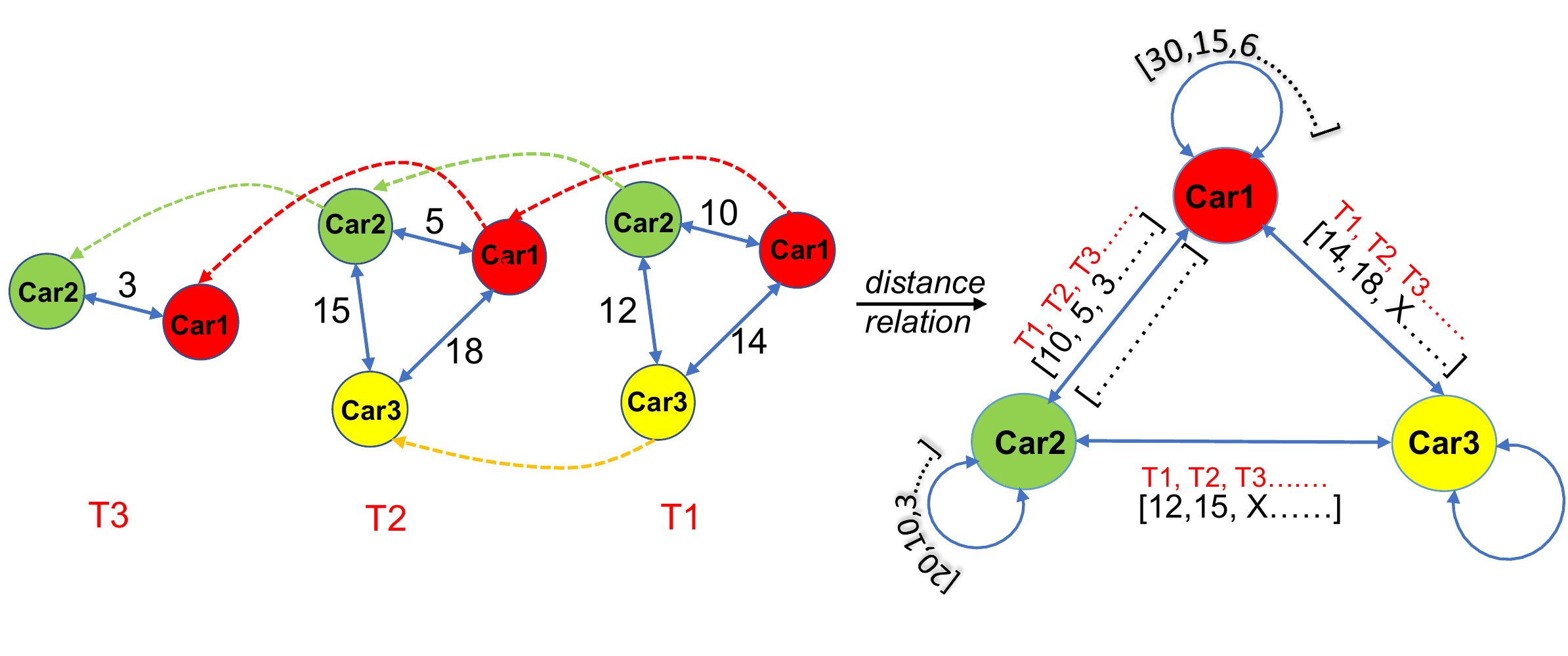}
\caption{VEKG-TAG construction from VEKG stream ~\cite{Yadav2019}}
\label{fig:8}
\end{figure}
\squeezeup

\subsubsection{Definition3 (VEKG-TAG):} For a given time $T$, having $n$ video frames represented as VEKG graph, the VEKG-Time Aggregated Graph is a labelled complete directed graph with 7 tuples such that $VEKG-TAG$ = $\{V,E,Av,R_E,T,$
\newline $\lambda_v,\lambda_E\}$. VEKG-TAG is similar to VEKG with an additional temporal dimension  $(T)$ adding to its edges in a single aggregated view (Algorithm \ref{algo:1}). It requires $\mathcal{O}(n^2T)$ memory to represent the VEKG stream of time T. Figure \ref{fig:8} shows a VEKG stream (left-side) and a VEKG-TAG (right-side) for time T1, T2, and T3 with a \textit{distance} relationship for three car objects. VEKG-TAG shows unique object nodes (car1, car2, and car3) and the \textit{distance} among them over time T1, T2, and T3. The distance between car1 and car2 decreases over time, which means car1 is approaching car2. The distance between car2 and car3 increases at T1 and T2, but since there is no car3 at time T3, it is represented by a don’t-care (X) condition. Each object node in VEKG-TAG has a \textit{self-loop} which stores its initial position for the image frame. This helps in capturing object dynamics such as the object is \textit{stationary} or \textit{moving} over time. VEKG-TAG consists of $[n(n-1)+n(self-loops)]$ edges which is equivalent to total $n^2$ edges.

\begin{spacing}{0.1}
\begin{algorithm}[H]
\caption{VEKG-TAG Algorithm}
\label{algo:1}
\SetAlgoLined
\textbf{Input:} \ Video Stream\

\KwResult{VEKG-Time Aggregated Graph }
 initialization\;
 \While{ video not null}{
  $frame_{i} \longleftarrow getframe(video)$\;
  
  $(object_{list},\ attribute_{list},\ bbox_{list}) \longleftarrow ModelCascade(frame_i)$\;
  
  $VEKG_{i} \longleftarrow Q(object_{list},attr_{list},bb_{list})$\;
  
  $win \longleftarrow addtowindow(VEKG_{i})$\;
  
  \eIf{win.size < trigger-time}{
  continue\;
  }{
   
  \For{$each VEKG_{i} \ in \ Win$ } {
   
  $V \longleftarrow V \cup (getnodes(VEKG_{i})$ \;
  }
   
  $VEKG-TAG \longleftarrow initializeTAG(V,E)$\;
   
  \For{$each VEKG_{i} \ in \ Win$ } {
   
  $VEKG-TAG \longleftarrow updateTAG(VEKG_{edge})$ \;
  }
   
  }
 }
\end{algorithm}
\end{spacing}

\section{System Architecture }
\label{S:9}

\begin{figure}[H]
\centering\includegraphics[width=10.1cm,height =4.6cm]{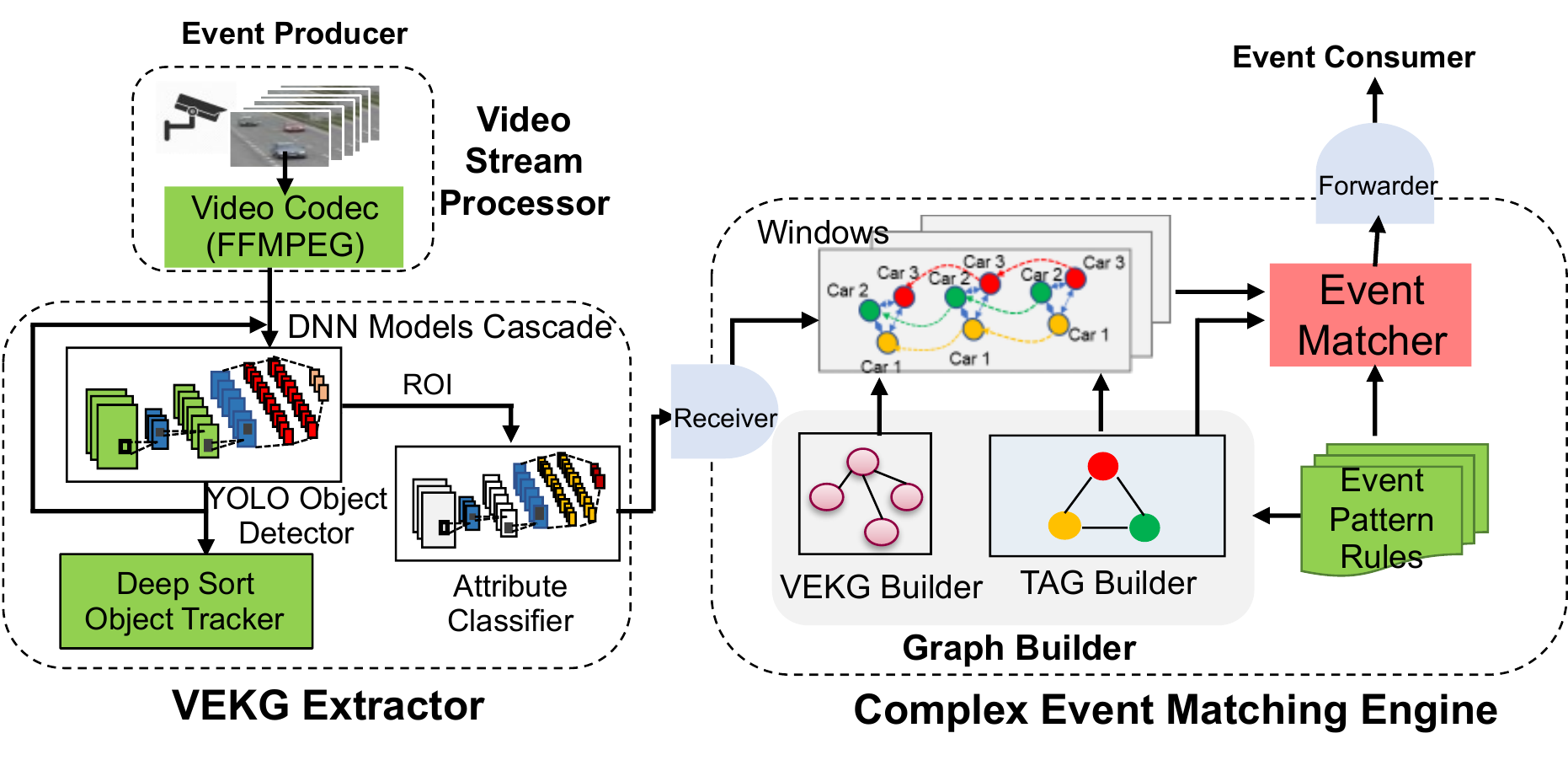}
\caption{VEKG extraction architecture}
\label{fig:9}
\end{figure}

Figure \ref{fig:9} shows the proposed architecture of VEKG extraction and  complex event pattern matching. The architecture is divided into three major components which are explained below in detail.

\squeezeup
\squeezeup

\begin{itemize}
    \item \textit{Video Stream Processor}: The \textit{video stream processor} receives the raw video frames from \textit{event producer} and processes them to low-level feature map using video encoders. The \textit{event producer} can be any streaming source generating video data such as CCTV camera, social media platforms (like YouTube), and web cameras. The video encoder (such as FFmpeg) receives the video data and convert them frame by frame to machine-readable features, which is further processed by the \textit{VEKG extractor} component.
    \item \textit{VEKG Extractor}: The \textit{VEKG extractor} receives the feature map of video frames and extract high-level semantic concepts such as objects and its attributes which are required to construct the VEKG graph. The VEKG extractor is a computer vision pipeline of different DNN models, such as object detectors, pose detectors, attribute classifiers, and trackers. Figure \ref{fig:9} shows a 3-stage model cascade with YOLO Object Detector~\cite{redmon2016you}, DeepSORT object tracker \cite{Wojke2017simple} and a attribute classifier. The object detector detects objects from the received feature map. The tracker keeps track of the object across frames so that unique objects can be identified. The features of the object are then passed to attribute classifiers to detect its characteristics such as \textit{color}. The model cascade complexity can vary depending on the requirements and use cases.
    
\item \textit{Complex Event Matching Engine}: The CEP matching engine has four components:
\subitem \textit{Event Rules}: The \textit{event rules} act as registry to store the rules for different patterns, which is used for VEKG graph creation and pattern matching.
\subitem \textit{Windows}: The \textit{windows} receive the VEKG stream as a \textit{state}. In this work, we are focusing on the time window (equation \ref{eq:1}).
\subitem \textit{Graph Builder}: The \textit{graph builder} receives the raw information from model cascades and creates VEKG graphs using \textit{event rules}. The graph builder consists of \textit{VEKG Builder}, which creates an initial VEKG stream over windows and a \textit{TAG Builder}, which aggregates the VEKG stream to create VEKG-TAG graph for a given window length. This is a multiprocessing system where VEKG creation and aggregation occurs in parallel.
\subitem \textit{Matcher}: The \textit{matcher} receives a VEKG stream from \textit{windows} or an aggregated VEKG-TAG from \textit{TAG builder}. As per the event rule, the matcher performs inference over VEKG (or VEKG-TAG) to identify whether a  pattern is satisfied or not.
\end{itemize}

\section{Spatiotemporal Relation for Video Event Knowledge Graph }
\label{S:10}

\squeezeup
\begin{figure}
\centering\includegraphics[width= 1\linewidth]{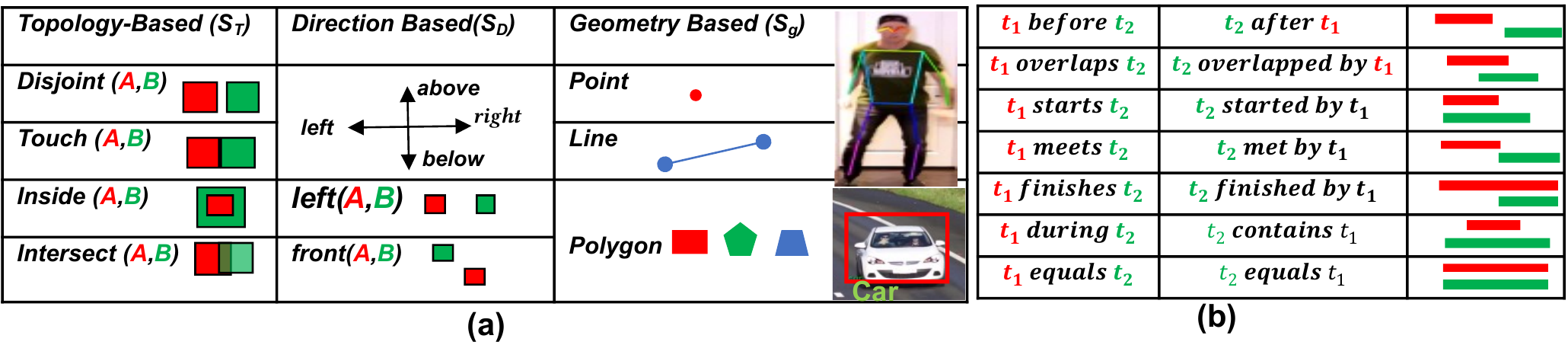}
\squeezeup
\caption{ (a) Spatial Relationships and (b) Temporal Relationships}
\label{fig:10}
\end{figure}
\squeezeup

Qualitative Spatial and Temporal Representation and Reasoning (QSTR) is a well-established field in artificial intelligence which relates to reasoning about space and time. QSTR provides formalism techniques such as conceptualization of space, spatiotemporal dynamics and common-sense reasoning to infer actions and changes in spatial and temporal dimensions~\cite{Cohn2001,Allen1981}. We have used spatiotemporal calculus to formalize and interlink semantic concepts across multimedia data streams which helps in building richer and queryable event patterns. One can develop intuitive event definitions, facilitating interactions between complex event defined by developer and domain expert.

\subsection{Spatial Relations}
\subsubsection{Geometric}
As shown in Figure \ref{fig:10}(a), a spatial entity can be represented using geometry-based features like point, line and polygon. The point represents the discrete features, while a line represents a linear feature. Similarly, a polygon represents a bounded region across some space. Semantic concepts like objects can be represented in different forms. It can be in terms of contour, segmented region, or a simple bounding box representation. The abstraction of an object depends on the granularity of the model, how it extracts, and represents an object. For example, an object can be described as a whole in bounding box or segmented region or parts like head, leg, and hand in a line based skeletal representation. We have used polygon-based bounding boxes and line-based skeletal representation for objects (figure \ref{fig:10}(a)).

\subsubsection{Topology}
We have used Dimensionally Extended nine-Intersection Model (DE-9im), a 2-dimensional topological model that describes pairwise relationships between spatial geometries (Sg). This mathematical model is based on Clementini Matrix~\cite{Clementini1993}, which describes relationships between geometries based on their interior(I), boundary(B), and exterior(E) features. The nine relationships it captures are- \{\textit{Disjoint, Touch, Contains, Intersect, Within, Covered by, Crosses, Overlap, Inside}\}. The four topological relations are shown in the (Figure \ref{fig:10}(a)) using a bounding box(polygon) based spatial geometry (Sg).

\subsubsection{Direction}
Direction captures the projection and orientation of an object in space. We have used Fixed Orientation Reference System (FORS)~\cite{Hernandez1991}, which divides the space into eight regions: \{\textit{above, below, left, right, left above, right above, left below, right below}\}. As shown in figure \ref{fig:10}(a), we took a simpler version of FORS with only four significant direction relations. There are also various qualitative models for orientation, which captures the angular relationship between objects. In our present work, we have considered only angular relations between objects, which are represented as lines.

\subsubsection{Spatial Builtin Functions} To perform spatial operations over objects we have devised two types of spatial operations:
\begin{itemize}
    \item \textit{Boolean Relation Operation}: This function returns boolean relation (0 and 1) between objects. For example the result for $Overlap(O_1, O_2)$ will be 1 if both object's bounding boxes are getting overlapped and vice versa. 
    \item \textit{Metric Relation Operation}: This function return relationship in numbers between objects. For example $DISTANCE(O_1, O_2)$ will return the distance between objects with respect to the image frame.
    
\end{itemize}

\subsection{Temporal and Logical Rules}
Since the above spatial interaction happens across time so we need to incorporate temporal relation during event matching. We have used the Allen 13 time-intervals~\cite{Allen1981} which helps in matching temporal patterns over VEKG stream (figure \ref{fig:10}). Except the spatial and temporal relation, we have used the logic operators $(AND (\wedge), OR (\vee) , NOT (\neg), ANY (\exists), EVERY (\forall), NOR (\downarrow) , XOR (\oplus), X-$
\newline$NOR(\odot),$ 
$Implies (\rightarrow ), Bi-Implies (\Leftrightarrow))$, mathematical and comparison operators $(+,-,*,$
$/, < >=)$ to model the relationships.

\section{Event Pattern Rules for Video Patterns in Different Domains }
\label{S:10}

To show the efficiency of VEKG, we have defined nine event pattern rules across two domains 1) Activity Recognition and 2)Traffic Management.

\squeezeup
\begin{figure}[H]
\centering\includegraphics[width= 1\linewidth]{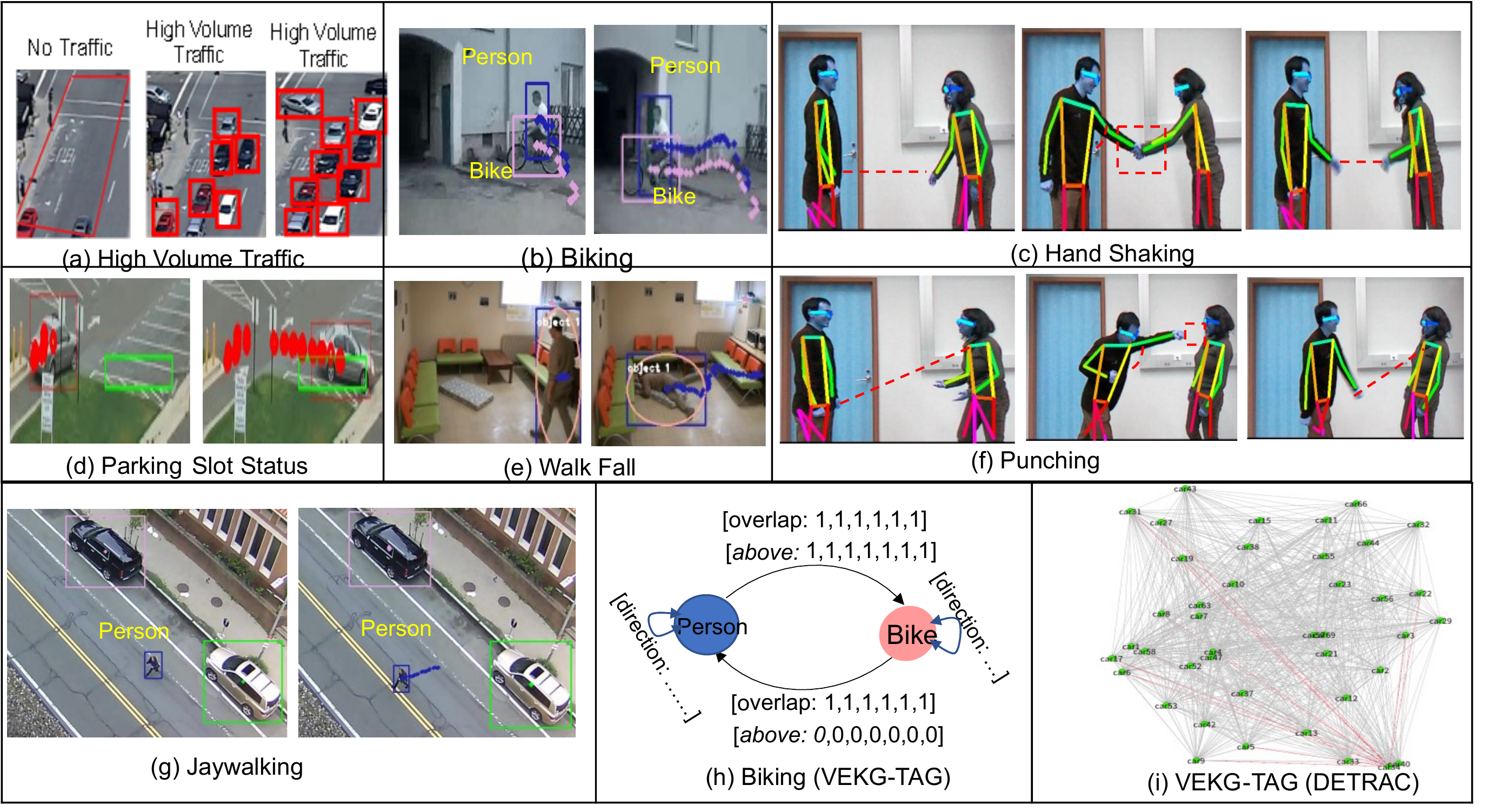}
\caption{Different event patterns and VEK-TAG example }
\label{fig:11}
\end{figure}
\squeezeup

\subsection{Activity Recognition}
In \textit{activity recognition}, we have defined 5 rules on the basis of a person's activity and action.

\subsubsection{Fall Detection}
Fall detection can be of multiple types depending on location and activity being performed by the person. In this work, fall detection rule is defined when a \textit{‘person falls while walking.’}. When a person falls, then there will be an abrupt change in its aspect ratio (here person is represented as a bounding box), and after that, there will be no motion of him for some time period (figure \ref{fig:11}(e)). The equation \ref{eq:2} modeled the \textit{fall detection} rule as:
\begin{equation}\scalebox{0.87}{$
    [ \ [AbruptChangePoint\left(AspectRatio(O)\right)]^{[t_i,t_j]} \ AND 
    [Motion\left(O\right)]^{[t_i,t_j] } = \alpha \ ]^{(\boxplus \ [t_1,t_2 ])}$}
\label{eq:2}
\end{equation}

where $O = person$, $\alpha = no \ motion$ and $\boxplus$ is a time window such that  $t_1 \leq t_i,t_j \leq t_2$. The aspect ratio of person is modeled as time-series data, and abrupt change point is detected using PELT algorithm~\cite{killick2012optimal}.
\subsubsection{Horse Riding}
The \textit{horse riding} rule is defined as:
\begin{equation}\scalebox{0.89}{$
    \left[ Overlap (O_1, O_2) \  AND \ Above(O_1, O_2) \ AND \ SameDirection(O_1, O_2) \right]^{(\boxplus \ [t_1,t_2 ])}$}
    \label{eq:3}
\end{equation}
where $O_1 = person$ and $O_2 = horse$. The horse riding rule ( equation \ref{eq:3}) states that the person bounding box should \textit{overlap} with horse bounding box such that person is \textit{above} the horse, and both are \textit{moving} in the \textit{same direction} for a given time window $t_1, t_2$.
\subsubsection{Bike Riding}
The \textit{bike riding} rule is similar to horse riding with the change in objects category where $O_1 = person$ and $O_2 = bike$. Figure \ref{fig:11}(b) shows a \textit{biking} event where a person object is \textit{above} bike and both are moving in the \textit{same} direction (blue and pink dots). Figure \ref{fig:11}(h) shows a VEKG-TAG for biking event where the edge from \textit{person} to \textit{bike} satisfies the \textit{overlap} and \textit{above} pattern. The edge weight consists of boolean values as boolean spatial operation is performed over these two objects. The direction relation (self-loop) stores the motion values of each object and can be compared during matching.

\subsubsection{Handshaking} A simple handshake can be viewed as two persons raise their hand to shake and later take the hand back to normal position. The  \textit{handshaking} event rule can be defined as \lq if the angle $(\theta_1, \theta_2)$ between two persons arm and shoulder increases with the decrease in their wrist distance $(\beta)$ initially, and later the angle$(\theta_1, \theta_2)$ decreases with an increase in wrist distance$(\beta)$ then it is a \textit{handshake} pattern (figure \ref{fig:11}(c)).\rq

\squeezeup
\begin{equation}\scalebox{0.83}{$
\left[ \ [\alpha_1(\theta_1, \theta_2, \beta)]^{[t_i,t_j]} \ AND \ [\alpha_2(\theta_1, \theta_2, \beta)]^{[t_{j+l},t_k]} \right]^{(\boxplus \ [t_1,t_2 ])} =
    \begin{cases}
      [t_i, t_j] & (\theta_1, \theta_2) \ increases\\
      [t_i, t_j] & (\beta) \ decreases\\
      [t_{j+l}, t_k] & (\theta_1, \theta_2) \ decreases\\
      [t_{j+l}, t_k] & (\beta) \ increases\\
    \end{cases}$}
\end{equation}
\begin{equation}\scalebox{0.85}{$
\begin{split}
    where \ \alpha = [ Angle(RightArm(O_1), RightShoulder(O_1)) = \theta_1 \ AND \\
    Angle(RightArm(O_2), RightShoulder(O_2)) = \theta_2 \ AND \\ Distance(RightWrist(O_1), RightWrist(O_2)) = \beta]
\end{split}$}
\label{eq:5}
\end{equation}
where $\theta_{1}$ and $\theta_{2}$ are an \textit{acute angle}, $(\beta)$ is the distance between the wrists and $O_{1},O_{2} \in  \ person$.

In equation \ref{eq:5}, the body parts like the right arm and shoulder can be replaced with left if there is a handshake from the left hand, as shown in the SBU-Kinetic dataset~\cite{kiwon_hau3d12} (figure \ref{fig:11}(c)).

\subsection{Punching} The punching event defined here is a \textit{single hand punching} (figure \ref{fig:11}(f)). The punching rule (equation \ref{eq:6}) can be written as \lq if there is an increase in the angle between person's arm and shoulder (acute or obtuse) and the distance between his wrist and another person's shoulder decrease in given time and later the angle decrease with increase in the distance.\rq
\begin{equation}\scalebox{0.89}{$
\left[ \ [\alpha_1(\theta, \ \beta)\ ]^{[t_i,t_j]} \ AND \ [\alpha_2(\theta, \beta)\ ]^{[t_{j+l},t_k]} \right]^{(\boxplus \ [t_1,t_2 ])} =
    \begin{cases}
      [t_i, t_j] & (\theta) \ increases\\
      [t_i, t_j] & (\beta) \ decreases\\
      [t_{j+l}, t_k] & (\theta) \ decreases\\
      [t_{j+l}, t_k] & (\beta) \ increases\\
    \end{cases}$}
    \label{eq:6}
\end{equation}
\begin{equation}\scalebox{0.89}{$
\begin{split}
    where \ \alpha = [ Angle(RightArm(O_1), RightShoulder(O_1)) = \theta_1 \ AND \ Distance \\(RightWrist(O_1) ,(RightShoulder(O_2)\  OR \ LeftShoulder(O_2))) = \beta \ ]
\end{split}$}
\label{eq:7}
\end{equation}

where $\theta_{1}$ can be \textit{acute or obtuse angle}, $(\beta)$ is the distance between the wrist and shoulder and $O_{1},O_{2} \in  \ person$.

\subsection{Traffic Management}
In traffic management we have defined four event rules which are as follows:
\subsubsection{High Volume Traffic} ‘High Volume Traffic’ rule (equation \ref{eq:8}) is defined as: ‘the average count of objects at a given space is greater than a certain threshold for a specific time range.’ For example, if the average number of cars is greater than 5 in every frame at a specific location of the road for more than 5 minutes, then we termed it as high volume traffic for that location (figure \ref{fig:11}(a)). It is defined as:

\begin{equation}
\exists \ \eta \in G \ and \ \forall t_i \in T \ if
\left(M(O)_\eta\right)^{(\boxplus \ [t_1,t_2 ])} =
    \begin{cases}
      $> \ r$ & \textit{traffic}\\
      $< \ r$ & \textit{not traffic}\\
    \end{cases}
    \label{eq:8}
\end{equation}

where $G$ is a \textit{space} and $T$ is \textit{time} such that $M=Avg.COUNT, \ O=car \ and \ r \in \mathbb{Z} $.

\subsubsection{Parking Slot Status}
We define a Parking slot status as \textit{occupied} ‘if the overlap of a queried object over a parking slot is greater than some threshold', then we can say that the object is occupying the parking slot (figure \ref{fig:11}(d)). The parking lot full pattern (equation \ref{eq:9}) can be written as:
\begin{equation}\scalebox{0.89}{$
\begin{split}
\exists \ \eta_{slot} \in G \ and \ \forall(O) \ at \ t_i  \in T \ if \ mro(ST(\eta_{slot},O))>r                   
\end{split}$}
\label{eq:9}
\end{equation}

where \textit{ST} = overlap \ is a spatial topological relation, \textit{mro} is \textit{metric relation operation}, $\eta_{slot}$ is parking slot, $\ O=car$ \ and $\ r \in \ real number$.

\subsubsection{Jaywalking}
The Jaywalking is defined as \lq if the person exists \textit{inside} the given road cross-section\rq, then he is Jaywalking (figure \ref{fig:11}(g)). Since the road cross-section does not frequently change so, we can give the dimensions of the required cross-section as a configuration parameter to identify the pattern. The Jaywalking rule (equation \ref{eq:10}) can be written as:  
\begin{equation}\scalebox{0.85}{$
\begin{split}
\exists \ \eta_{road \ cross \ section} \in G \ at \ t_i  \in T \ if \ msf(ST(\eta_{road \ cross \ section},O)) = True  \ where \\
ST=Inside \ and \ O=person                
\end{split}$}
\label{eq:10}
\end{equation}

\subsubsection{Car With Specific Attributes}
During traffic management, the traffic authority may be interested in finding objects (like a car) with specific attributes such as color and license plate number. The rules for such events (equation \ref{eq:11}) can be written as:

\begin{equation}
    Detect(object) \ where  \ object = car \ AND  \ object.color \ = \ 'Red'
\label{eq:11}
\end{equation}
Detecting such events will require multi-stage DNN models where it can detect the objects and their attributes.

Figure \ref{fig:11}(i) shows a complex VEKG-TAG over DETRAC~\cite{DETRAC:CoRR:WenDCLCQLYL15} dataset where nodes are different \textit{car} objects which exists over a given time window. The edges (like red color) can be directly fetched between object nodes and inference can be performed to identify whether a pattern exists or not.

\section{Experiments and Results }
\label{S:11}
\subsection{Implementation and Datasets}
The prototype of the system is implemented in Python 3 over VidCEP~\cite{yadav2019vidcep,yadav2019high} engine. The experiments were performed over a Linux machine with 16 core Intel® i9-9900K CPU, 64 GB RAM, and Nvidia GeForce RTX 2080 Ti GPU. OpenCV~\cite{OpenCV} was used for initial video processing. The model cascade used a pre-trained (MSCOCO) YOLOv3~\cite{redmon2016you} object detector, Posenet~\cite{papandreou2018personlab} for pose detection and DeepSORT~\cite{Wojke2017simple} for object tracking. The attribute classifier is a color filter implemented in OpenCV. The features of bounding boxes extracted from object detectors are passed to attribute classifiers for detecting an object's characteristics (here color). The NetworkX~\cite{NetworkX} python library was used for creating VEKG graphs. The experiments were performed across 10 datasets over 801 videos related to different event patterns.  Table \ref{table:2} shows the list of datasets with the number of videos being processed for the corresponding event pattern. The number of videos for activity recognition is high, but they are small clips, while traffic management videos are less in number and have a long time duration.

\squeezeup
\begin{table}[H]
\begin{minipage}[b]{0.56\linewidth}
\caption{Dataset Specification}
\begin{adjustbox}{width=6.7cm, height=2.1cm,center}
\begin{tabular}{ | l | r | r |}
    \hline
    \textbf{Datasets} & \textbf{Total Videos}  & \textbf{Event Pattern} \\
    \hline
    \hline
    \textbf{L2ei}~\cite{charfi2013optimized} & 213 & Fall Detection \\
    \hline
        \textbf{MuHAVi-MHI}~\cite{singh2010muhavi} & 14 & Fall Detection \\
    \hline
        \textbf{HMDB}~\cite{Kuehne11} & 215 & Horse,Bike Ride \\
    \hline
        \textbf{UCF-101}~\cite{soomro2012ucf101} & 267 & Horse,Bike Ride \\
    \hline
        \textbf{UT-Interaction}~\cite{UT-Interaction-Data} & 60 & Handshaking,Punching \\
    \hline
            \textbf{SBU Kinetic}~\cite{kiwon_hau3d12} & 10 & Handshaking,Punching \\
    \hline
            \textbf{DETRAC}~\cite{DETRAC:CoRR:WenDCLCQLYL15} & 10 & High Volume Traffic \\
    \hline
            \textbf{Street Scene}~\cite{MERL_TR2018-188} & 5 & Jaywalking \\
    \hline
            \textbf{VIRAT}~\cite{oh2011large} & 5 & Parking Lot Status \\
    \hline
                \textbf{PEXELS}~\cite{Pexels} & 2 & Car Attribute \\
    \hline
            \textbf{Total Videos} & 801 & NA  \\
    \hline
  \end{tabular}
  \end{adjustbox}
    \label{table:2}
\end{minipage}\hfill
\begin{minipage}[b]{0.4\linewidth}
\caption{Event Accuracy Formula}
\begin{adjustbox}{width=5.7cm, height=1.7cm,center}
\centering
\begin{tabular}{ | l | }
    \hline
       \centerline{\textbf{\ \ \ Event Accuracy}} \\
        \hline
        \textit{True Positive}: relevant events detected correctly \\
            \hline
        \textit{False Positive}: events which are detected as relevant  \\
            \hline
        \textit{False Negative}: relevant events not detected \\
            \hline
        \textit{Precision} = $\dfrac{True Positive}{True Positive + False Positive}$ \\
    \hline
    \textit{Recall} = $\dfrac{True Positive}{True Positive + False Negative}$ \\
    \hline
    \textit{F-Score} = $\dfrac{2*Precision*Recall}{Precision+Recall}$\\
    \hline
  \end{tabular}
  \end{adjustbox}
    \label{table:3}
\end{minipage}
\end{table}
\squeezeup

\subsection{Event Accuracy}
The event accuracy is defined as how many relevant event patterns were detected for a given event rule as compared to the ground truth. Event accuracy is evaluated using F-Score (Table \ref{table:3}), which is a harmonic mean of \textit{precision} and \textit{recall}. The precision is the ratio of \textit{relevant events matched} and \textit{matched events} while recall is the ratio of \textit{relevant events matched} and \textit{relevant events}. Table \ref{table:4} shows the mean Precision, Recall, and F-Score for different event rules across multiple datasets and is compared with other state-of-the-art methods.

\begin{figure}[H]
\centering\includegraphics[width=1\linewidth]{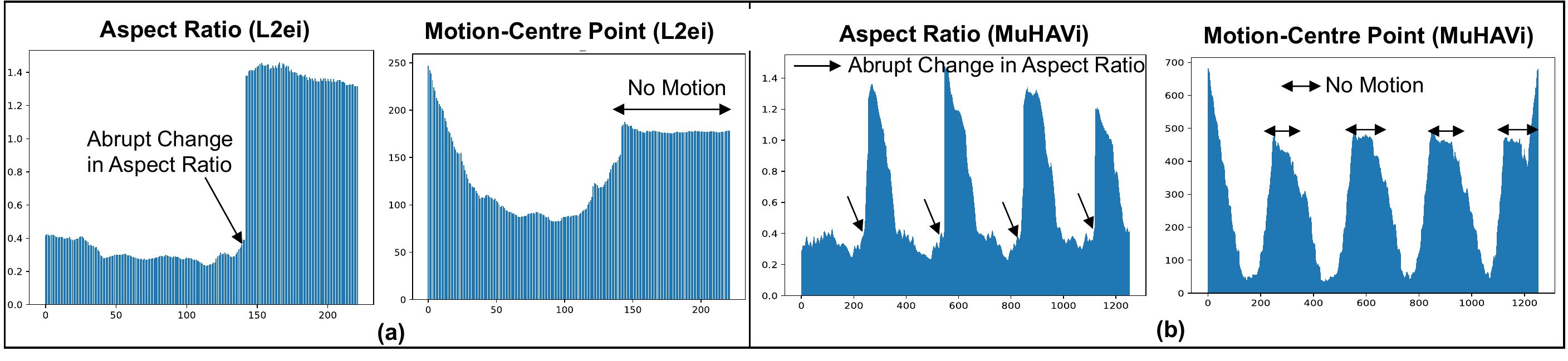}

\caption{Abrupt change in aspect ratio and motion during Fall detection in (a) L2ei and (b) MuHAVi dataset}
\label{fig:12}
\end{figure}

The \textit{Fall Detection} event rule was evaluated over two datasets- L2ei~\cite{charfi2013optimized} and MuHAVi-MHI~\cite{singh2010muhavi}. The VEKG F-Score for L2ei was 0.87 as compared to the SVM+STHF (filter)\cite{charfi2013optimized} method, which has an F-Score of 0.95. The SVM+STHF (filter) method used a prefiltering approach with a combination of different features to train the model to achieve such high accuracy. The F-Score of Fall detection over MuHAVi was 0.83. Figure \ref{fig:12} (a) and (b) shows two instances of fall detection in the L2ei and MuHAVi dataset. In figure \ref{fig:12} an abrupt change can be seen in the aspect ratio in both images, and after that, no motion of the person was detected. In L2ei (figure \ref{fig:12}(a)), there was a single fall, while in MuHAVi (figure \ref{fig:12}(b)), a person falls four times (four abrupt changes in the VEKG), which VEKG was able to identify correctly. In L2ei and MuHAVi dataset, only a single person is present, so all the information regarding motion and aspect ratio was saved in the self-loop edge of the person object node. Later the data in the edge is modeled as time series to identify the abrupt change and motion distribution using the PELT method~\cite{killick2012optimal}. The \textit{Horse Ride} event rule for HMDB~\cite{Kuehne11} and UCF-101~\cite{soomro2012ucf101} does not perform well and has a low F-Score of 0.44 and 0.52 respectively. The HMDB and UCF-101 datasets are complex, where clips are small and from movies that have multiple objects with different Field of View (FoV). The YOLO and DeepSORT tracking performance is not good for these datasets, and there were many false positives during pattern matching. The \textit{Bike Ride} rule for HMDB performs equivalent to HOG/HOF\cite{Kuehne11} method with F-Score of 0.84 and have an F-Score of 0.661 for UCF-101.  

The \textit{Handshaking} rule in UT-Interaction~\cite{UT-Interaction-Data} dataset achieves an F-Score of 0.88 as compared to ExtCore9+SVM~\cite{kalita2018efficient} method, which has an accuracy of F-Score 0.95. The reason for small F-Score in VEKG is because there are multiple different actions within the same video, which increases the false positives. The other reason is the errors from the Posenet model, where it was unable to detect the correct pose orientation. VEKG outperforms ExtCore9+Naive Bayes method~\cite{kalita2018efficient} (F-Score - 0.46) and achieves an F-Score of 0.769 for \textit{Handshaking} pattern on SBU-Kinetic~\cite{kiwon_hau3d12} dataset. The excellent accuracy is due to the reason that there are only two persons involved in activity with the right Field of View (FoV). The VEKG method on the \textit{Punching} event on UT-interaction achieves F-score 0.518 and gives better results as compare to ExtCore9+Naive Bayes, which has an F-Score of 0.5.
Similarly, the VEKG achieves 0.66 F-Score on SBU-Kinetic for the Punching event. The VEKG achieves the highest F-Score (0.90) for the \textit{High Volume Traffic} event on the DETRAC~\cite{DETRAC:CoRR:WenDCLCQLYL15} dataset because it counts object nodes and depends on the accuracy of YOLO object detector. The \textit{Parking Slot Status} pattern on the VIRAT~\cite{oh2011large} dataset has the  F-Score of 0.79 as there were instances where a car bounding box was getting overlapped to multiple parking lots bounding boxes due to incorrect FoV.  The \textit{Jaywalking} on Street Scene~\cite{MERL_TR2018-188} dataset has F-Score of 0.89, where it was able to detect person on the street. The F-Score of the \textit{Car Attribute} query was 0.78 because of the low accuracy of the attribute classifier.

\squeezeup
\squeezeup

\begin{table}[H]
    \centering
    \caption{Event Accuracy comparison with sate-of-the-art method}
\begin{adjustbox}{width=\columnwidth, height=2.8in,center}
\begin{tabular}{ |c|c|c|c|c|c| }
\hline
Event Pattern & Datasets & Methods & Precision & Recall & F-score \\
\hline
\hline
\multirow{5}{4em}{Fall Detection} & \multirow{5}{4em}{L2ei} & SVM+STHF \ (no \ filter)~\cite{charfi2013optimized} & 0.94 & 0.921 & 0.929\\
               &      & SVM+STHF \ (filter)~\cite{charfi2013optimized} & 0.942 & 0.980 & 0.959 \\
               &      & Adaboost+STHF \ (no \ filter)~\cite{charfi2013optimized} & 0.884 & 0.901  & 0.88 \\
               &      & Adaboost+STHF \ (filter)~\cite{charfi2013optimized} & 0.951 & 0.921  & 0.93 \\
               &      & \textbf{VEKG} & \ 0.828

 & 0.915 & \textbf{0.869} \\
\hline
\multirow{4}{4em}{Fall Detection} & \multirow{4}{4em}{MuHAVi MHI \ } & LOCO~\cite{murtaza2016multi} & 0.16 & 0.014 & 0.025\\
               &      & LOAO~\cite{murtaza2016multi} & 0.893 & 0.952 & 0.921 \\
               &      & LOSO~\cite{murtaza2016multi} & 0.98 & 1 & 0.99 \\
               &      & \textbf{VEKG} & 0.785 & 0.88 & \textbf{0.83} \\
\hline
\multirow{3}{4em}{Bike Ride} & \multirow{3}{4em}{HMDB } & HOG/HOF~\cite{Kuehne11} & 0.823& 0.875 & 0.848\\
               &      & C2~\cite{Kuehne11} & 0.7142	& 0.769 & 0.740 \\
               &      & \textbf{VEKG} & 0.727 & 1 & \textbf{0.84} \\
\hline
Bike Ride & UCF-101 & \textbf{VEKG} & 0.529 & 0.88 & \textbf{0.661}  \\
\hline
\multirow{3}{4em}{Horse Ride} & \multirow{3}{4em}{HMDB} & HOG/HOF~\cite{Kuehne11} & 0.857 & 0.666 & 0.75\\
               &      & C2~\cite{Kuehne11} & 0.6 & 0.75 & 0.66 \\
               &      & \textbf{VEKG} & 0.325 & 0.8 & \textbf{0.44} \\
\hline
Horse Ride & UCF-101 & \textbf{VEKG} & 0.46 & 0.59 & \textbf{0.52} \\
\hline
\multirow{5}{4em}{Hand Shaking} & \multirow{5}{4em}{UT Interaction} & ExtCORE9 \ + \ KNN~\cite{kalita2018efficient} & 0.9 & 0.9 & 0.9\\
               &      & ExtCORE9+SVM~\cite{kalita2018efficient} & 1 & 0.9 & 0.95 \\
               &      & ExtCORE9+Naive Bayes~\cite{kalita2018efficient} & 1 & 1 & 1 \\
               &      & ExtCORE9+DeepLearning~\cite{kalita2018efficient} & 1 & 1 & 1 \\
               &      & \textbf{VEKG} & 0.857

 & 0.923 & \textbf{0.880} \\
\hline
\multirow{5}{4em}{Hand Shaking} & \multirow{5}{4em}{SBU Kinetic} & ExtCORE9 \ + \ KNN~\cite{kalita2018efficient}  & 0.44 & 0.42 & 0.43\\
               &      & ExtCORE9+SVM~\cite{kalita2018efficient} & 0.39 & 0.47 & 0.43 \\
               &      & ExtCORE9+Naive Bayes~\cite{kalita2018efficient} & 0.45 & 0.47 & 0.46 \\
               &      & ExtCORE9+DeepLearning~\cite{kalita2018efficient} & 0.44 & 0.4 & 0.42 \\
               &      & \textbf{VEKG} & 0.714
 & 0.833
 & \textbf{0.769} \\
\hline
\multirow{5}{4em}{Punching} & \multirow{5}{4em}{UT Interaction} & ExtCORE9 \ + \ KNN~\cite{kalita2018efficient}  & 0.39 & 0.5 & 0.44\\
               &      & ExtCORE9+SVM~\cite{kalita2018efficient} & 0.5 & 0.2 & 0.3 \\
               &      & ExtCORE9+Naive Bayes~\cite{kalita2018efficient} & 0.43 & 0.6 & 0.5 \\
               &      & ExtCORE9+DeepLearning~\cite{kalita2018efficient} & 0.33 & 0.5 & 0.4 \\
               &      & \textbf{VEKG} & 0.636 & 0.437 & \textbf{0.518}
 \\
\hline
\multirow{5}{4em}{Punching} & \multirow{5}{4em}{SBU Kinetic} & ExtCORE9 \ + \ KNN~\cite{kalita2018efficient} & 0.41 & 0.39 & 0.4\\
               &      & ExtCORE9+SVM~\cite{kalita2018efficient} & 0.92 & 0.61 & 0.73 \\
               &      & ExtCORE9+Naive Bayes~\cite{kalita2018efficient} & 0.86 & 0.67 & 0.75 \\
               &      & ExtCORE9+DeepLearning~\cite{kalita2018efficient} & 0.59 & 0.56 & 0.57 \\
               &      & \textbf{VEKG} & 0.571
 & 0.8 & \textbf{0.66} \\
\hline
High Volume Traffic & DETRAC & \textbf{VEKG} & 0.91 & 0.89 & \textbf{0.90} \\
\hline
Parking Lot Status & VIRAT & \textbf{VEKG} & 0.83 & 0.75 & \textbf{0.79} \\
\hline
Jaywalking & StreetScene & \textbf{VEKG} & 0.86 & 0.92 & \textbf{0.89} \\
\hline
Car Attribute & PEXELS & \textbf{VEKG} & 0.81 & 0.76 & \textbf{0.78} \\
\hline
\end{tabular}
\end{adjustbox}
    \label{table:4}
\end{table}
\squeezeup

\subsection{VEKG and VEKG-TAG Metrics}

\subsubsection{Reduction in Nodes and Edges}
The efficacy of VEKG-TAG can be measured in terms of the number of nodes and edges being reduced for the matching. VEKG-TAG reduce the redundant nodes and edges from the VEKG stream. Reduction in Nodes (RIN) is defined as the ratio of the difference in the number of nodes between VEKG stream $(|VEKG_{nodes}|)$ and VEKG-TAG graph $(|VEKG-TAG_{nodes}|)$  with the number of VEKG nodes (equation ~\ref{eq:12}). The Reduction in Edge (RIE) follows the same pattern where nodes are replaced with edges between both graphs to measure its effectiveness (equation ~\ref{eq:13}).

\begin{minipage}{.44\textwidth}
\begin{equation}\scalebox{0.77}{$
RIN = \dfrac{|VEKG_{nodes}| - |TAG_{nodes}|}{|VEKG_{nodes}|} $} 
\label{eq:12}
\end{equation}
\end{minipage}%
\begin{minipage}{.44\textwidth}
\begin{equation}\scalebox{0.77}{$
RIE = \dfrac{|VEKG_{edges}| - |TAG_{edges}|}{|VEKG_{edges}|} $}
\label{eq:13}
\end{equation}
\end{minipage}
\vskip.8\baselineskip

\begin{figure}[H]
\centering\includegraphics[width=1\linewidth]{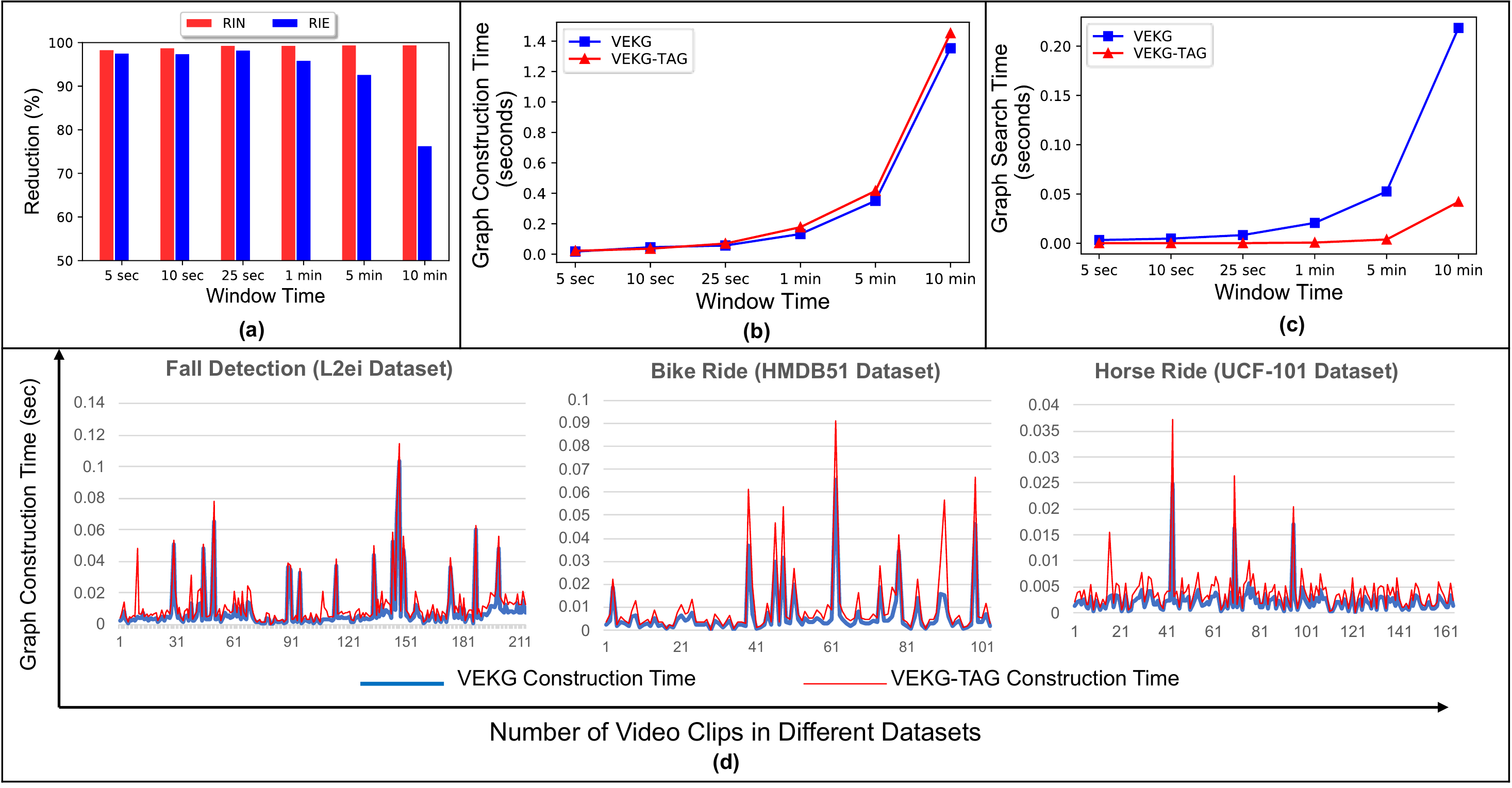}
\caption{VEKG Metrics for different window size (a) Reduction in nodes(RIN) and Edges(RIE), (b) VEKG and VEKG-TAG construction time, (c) VEKG and VEKG-TAG search time and (d) VEKG and VEKG-TAG construction time for three datasets}
\label{fig:13}
\end{figure}

Figure \ref{fig:13}(a) shows the RIN and RIE score over Street Scene dataset for different windows ranging from 5 seconds to 10 minutes. The windows size of 10 minutes has approximately 18000 frames for 30 frames per second (fps) video stream. In such a scenario, the number of objects will be very high, leading to an explosion in VEKG nodes. In Fig \ref{fig:13}(a), the RIN and RIE score for the 10-minute window was 99.5\% and 76.3\%. This means that VEKG-TAG reduces nearly 99\% of nodes and 76.3\% of edges from VEKG stream. The reduction in the number of nodes in Street Scene dataset is due to high number of stationary objects (such as car parked in the parking area) while the number of new incoming objects (new cars and people moving on the street) are less. In the given experiment, the VEKG-TAG reduces an average of 13563.3 nodes (99\%) and 58822.6 edges (93\%) across all window times.

\squeezeup
\squeezeup

\subsubsection{Graph Construction and Search Time}
The graph construction is the time to create a VEKG graph over a given time window. Graph construction includes the time for creating nodes and edges relations as per event pattern rule. Figure \ref{fig:13}(d) shows the VEKG and VEKG-TAG construction time over three datasets (L2ei, HMDB51, and UCF-101) for 485 videos. The construction time for VEKG and VEKG-TAG is nearly same for each video. There is an average 3.8-millisecond increase in VEKG-TAG construction time over VEKG across different videos in datasets. This sub-second increase is due to the extra time required by VEKG-TAG to initialize its nodes and edges and get label from VEKG. Figure \ref{fig:13}(b) shows the graph construction time for VEKG and VEKG-TAG for different time windows over Street Scene dataset. The graph construction time increases with the increase in window size as there will be more objects creating more nodes. For VEKG and VEKG-TAG, the construction time for a 5-second window was 0.017 seconds and 0.022 seconds, which increases to 1.35 and 1.45 seconds respectively for 10-minutes window. There was a 7.4\% increase in the construction time of VEKG-TAG as compare to VEKG.

The graph search time is the time to search the pattern as per the event rule. Figure \ref{fig:13}(c) shows the search time for both VEKG and VEKG-TAG methods for different window sizes. VEKG-TAG performs better in search as it is the summarized version of VEKG with non-redundant nodes and edges. For 5-second window, the search time of VEKG and VEKG-TAG is 0.003 and 0.0001 second, respectively. For a 10-minute time window, the VEKG-TAG search requires only 0.042 seconds as compared to VEKG, which has a search time of 0.218 seconds. Thus, the VEKG-TAG construction time was 1.07 times of VEKG, but its search time was 5.19 X faster for a 10-minute time window. The performance of VEKG-TAG will increase with the increase in window size and number of event rules. The performance shown here is under a worst-case scenario where all the nodes and edges were traversed for both graph methods.       

\squeezeup
\begin{figure}[H]
\centering\includegraphics[width=1\linewidth]{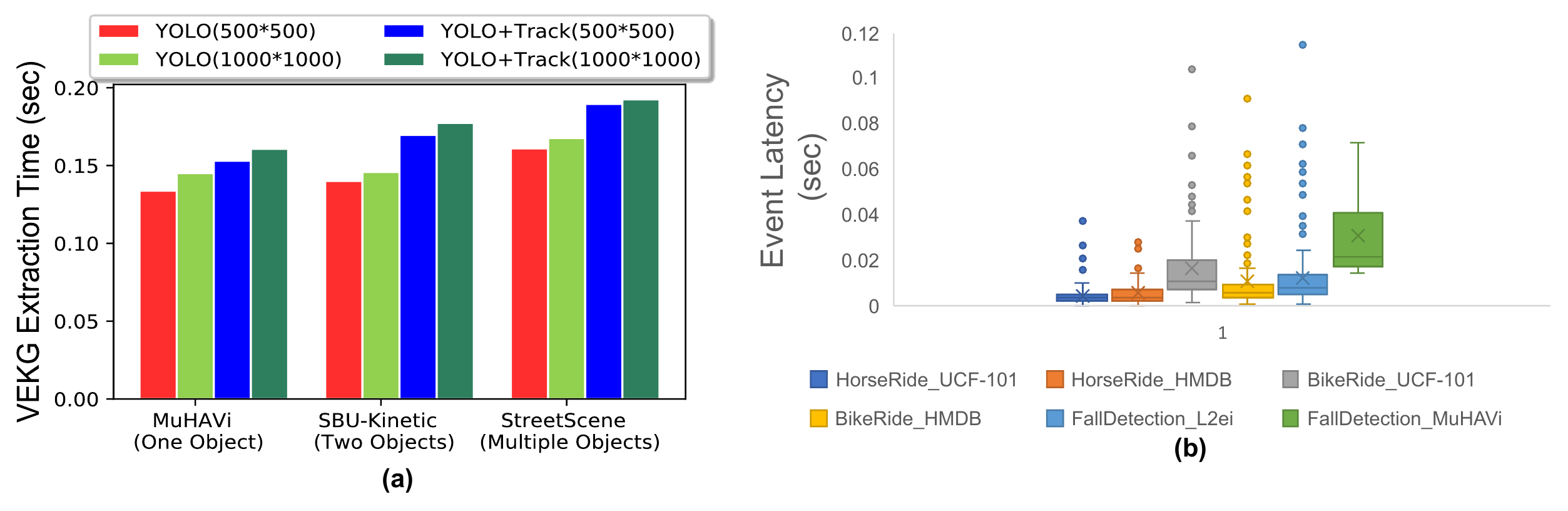}
\caption{ (a) VEKG extraction time and (b) Event Latency}
\label{fig:14}
\end{figure}
\squeezeup

\subsection{VEKG Extraction Time}
VEKG extraction is the total time required to process each video frame. Equation \ref{eq:14} shows the VEKG extraction time that includes the time to read the video frame from the encoder ($t_{video-encoder}$) and the DNN models inference time ($t_{Model  Cascade  Inference}$) to extract out the list of objects and its attributes.
\begin{equation}
    t_{VEKG-extraction} = t_{video-encoder} + t_{Model  Cascade  Inference}  
\label{eq:14}
\end{equation}

Figure \ref{fig:14}(a) shows the VEKG extraction time on three dimensions:
\begin{itemize}
\item \textit{Number of objects in frames}: The extraction time will be high if the number of objects is more in a frame. Although this time is very less due to the shared computation principle used by object detectors~\cite{redmon2016you} but in the long run, it affects the overall system performance. 
\item \textit{Number of models in a cascade}: The number of models directly affects the extraction time as now the frames need to pass to all different models to extract objects and their features.
\item \textit{Frame resolution}: The resolution directly affects the extraction time as the models need to process more features for high-resolution video frames.
\end{itemize}

In Figure \ref{fig:14}(a), the average event extraction time for one object (MuHAVi), was 0.133 seconds (500*500 resolution), which increases to 0.161 seconds for multiple objects (Street Scene) when processed over YOLO model. The event extraction time for 500*500 resolution frame increases from 0.133 seconds to 0.153 seconds when processed by a 2-stage model cascade (YOLO + DeepSort Tracking). There is a 43.3\% increase in the VEKG  extraction time between a single object 500*500 resolution frame (YOLO only) as compare to multiple objects 1000*1000 resolution frame when processed over a two-stage model cascade. Thus, VEKG extraction time is one of the bottlenecks, and its optimization will be the focus area of our future work. 

\subsection{Event Pattern Latency}
The event pattern latency is the time taken by the event rule to process the VEKG-TAG graph to detect a pattern. The event latency includes the time to apply rules over VEKG edges to create a relationship between object nodes $(VEKG_{construction})$, the time to create the VEKG-TAG $(TAG_{construction})$ and the search time to detect the pattern between nodes $(TAG_{search})$ (equation \ref{eq:15}).

\squeezeup
\begin{equation}\scalebox{0.89}{$
    Event-Pattern_{latency} = VEKG_{construction} + TAG_{construction}+ TAG_{search} $}
\label{eq:15}
\end{equation}

Figure \ref{fig:14}(b) shows the event latency time for three event rules (Bike Ride, Horse Ride, and Fall Detection) for six datasets. The latency is calculated over a time window of size equal to the length of the video clip. The figure shows the latency distribution time in a box plot for  videos related to the event pattern. The minimum median latency was 4 milliseconds for $Horse Ride-UCF-101$ dataset. The similar latency distribution is for the \textit{Bike Ride} datasets as both patterns same require topological \textit{(overlap, above)} operations. The processing also depends on the number of object nodes present in the video, which directly affects its search and construction time. The \textit{Fall Detection} rule for the MuHAVi dataset has the highest median latency of 20 milliseconds. Thus, VEKG is highly efficient in detecting video event patterns with sub-second latency.

\subsubsection{Limitations} The event rules are dependent on the DNN models performance for their reasoning. The prediction failure in DNN models leads to incorrect pattern identification.  Videos are complex, and writing generalized event rules is tricky. Presently, the pattern calculations are performed in a 2-D plane, which limits the expressiveness of event rules leading to many patterns misses or false event detection. In the real world, relations are quite complex and spread in 3-dimensions. The rules will not be efficient in cases of blind spots, the inconsistent field of view, and moving camera. Some events are highly complicated (such as Cooking and Juggling) and require deep feature learning. The present event pattern rules are more related to spatiotemporal nature, and complex models can be embedded in the CEP system to reason for more complex patterns.

\section{Conclusion and Future Work }
\label{S:12}
The work proposes a semantic representation technique to detect event patterns from unstructured video in CEP systems. We present the Video Event Knowledge Graph (VEKG), a graph driven approach to represent video streams and enables the CEP system to detect visual patterns. The work discusses the challenges and requirements which current CEP system requires to match video event patterns. The paper details the design and architecture for VEKG extraction. Later, VEKG- TAG is proposed, which is a state-based optimization over VEKG streams. The paper proposed a set of nine event rules from activity recognition and traffic management domain. The experiments were performed across 10 datasets having 801 videos and compared with state-of-the-art methods. The proposed approach achieves an F-Score ranging from 0.44 -0.90 with a 5.19X faster search time and sub-second matching latency. In the future work, we would focus on semantic enrichment of the VEKG and VEKG-TAG graphs to identify event patterns at different hierarchies, such as the relationship between identified object and geospatial location. We would improve the reasoning techniques by enriching depth data(3-D) to increase the overall pattern detection capability.



\section*{Acknowledgment}
This work was supported with the financial support of the Science Foundation Ireland grant SFI/13/RC/2094 and SFI/12/RC/2289\_P2. We also thanks Dibya Prakash Das from Indian Institute of Technology(IIT) Kharagpur for his initial contribution as an intern during the project.

\bibliographystyle{splncs04}
\bibliography{journal_v1}

\end{document}